# Does a completed simulation run mean a usable urban wind result for AI agents?

**Project Page:** https://ttanddd.github.io/urbanwind-agentbench/

**Zhongke Qu[1,†], Yunwei Zhang[1], Junwei Su[1], Yabin Zhang[2], Chengwei Li[3], Qingxiang Shui[4], Hailong Qu[5,*], Zhaolin Gu[1,*]**

[1]School of Human Settlements and Civil Engineering, Xi'an Jiaotong University, Xi'an 710049, China

[2]Shaanxi Institute of Meteorological Science,Xi'an, China

[3]Shaanxi Provincial Atmospheric Observation Technical Support Center, Xi'an,China

[4]Xihua University, School of Energy and Power Engineering

[5]School of Civil Engineering, Chongqing University, Chongqing, 400045, China

[†]These authors contributed equally to this article.

***Corresponding authors:**

Hailong Qu, School of Civil Engineering, Chongqing University, Chongqing, 400045, China.Email: tiyudi789520@gmail.com

Zhaolin Gu, School of Human Settlements and Civil Engineering, Xi'an Jiaotong University, Xi'an 710049, China.Email: guzhaoln@mail.xjtu.edu.cn

**Abstract**

AI agents based on large language models (LLMs) can now configure and run computational fluid dynamics (CFD) solvers, but a completed run does not ensure that delivered metrics can be recomputed from raw outputs. This gap was examined for urban wind simulation, where multi-case comparisons must hold non-target conditions fixed and reported values must be traceable to raw samples. UrbanWind-AgentBench, comprising 200 tasks and 367 cases in 10 engineering task families, was constructed. Public task contracts specify variable factors, fixed conditions, numerical thresholds and deliverable requirements, and two-layer scoring separates a structural score for deliverables, run provenance and raw evidence from a numerical score obtained by independent recomputation against references. WindPilot was then proposed as an execution framework that incorporates contract parsing, execution provenance, deterministic verification and bounded acceptance review into the LLM execution loop. On 30 OpenFOAM tasks with independent references, at least the structural maximum of 60 points was reached by a base agent on 28 tasks, but numerical credit was obtained on only 6. With the same model name and budget, WindPilot increased the number of tasks with numerical credit to 17, full-score tasks from 0 to 3 and the mean task score from 62.5 to 72.5. Mean action steps fell from 130.4 to 104.9, and mean token consumption decreased by 12.9%. The mean gain for atmospheric boundary layer (ABL) tasks was 18.5 points. A completed simulation run is therefore not sufficient for a usable urban wind result; usability depends on checks that tie reported metrics to raw evidence during execution.

## Graphical abstract

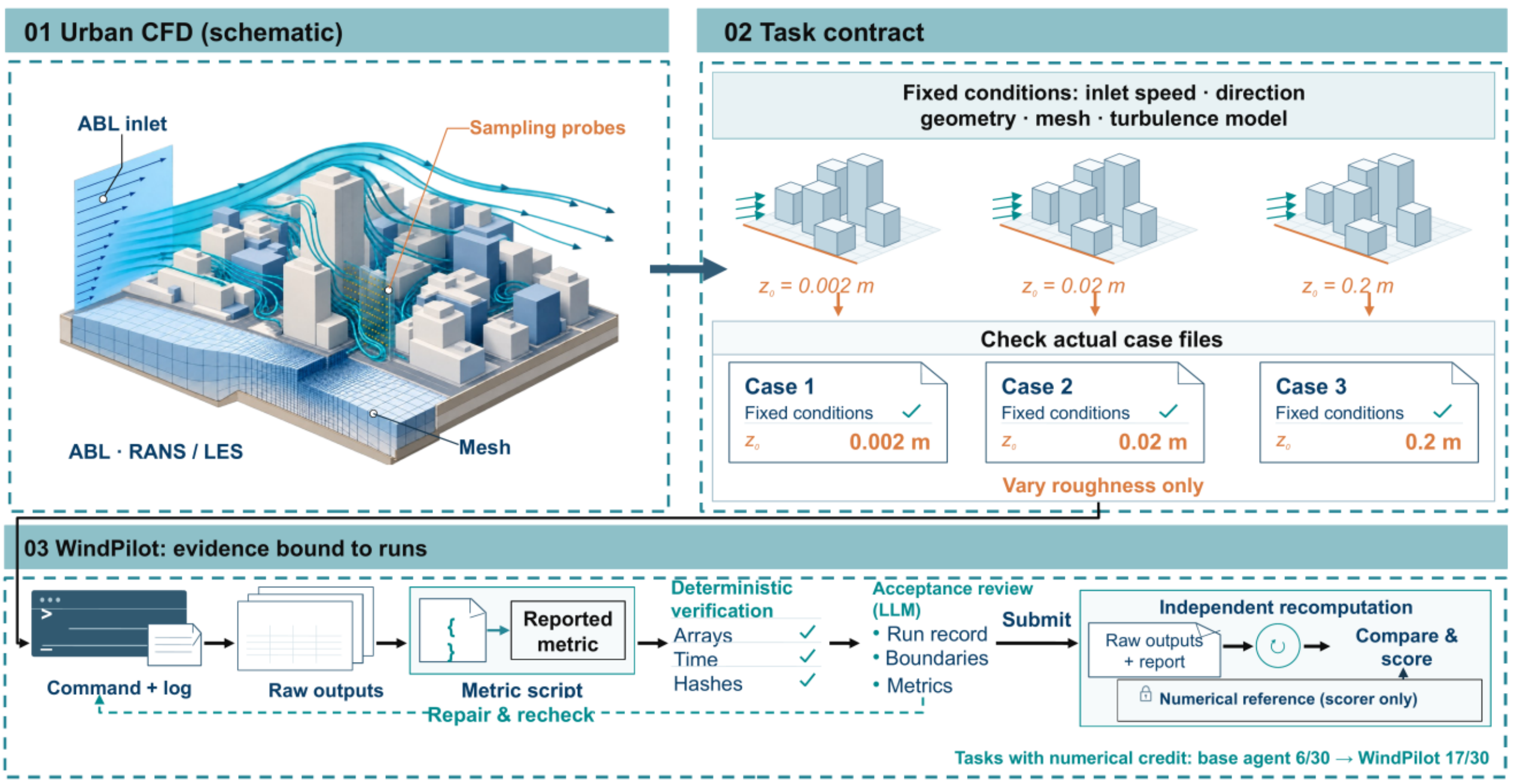

# 1 Introduction

Urban wind assessment supports decisions about pedestrian comfort, natural ventilation and pollutant dispersion, and computational fluid dynamics (CFD) is the main method for such assessment [1,2]. Reliable assessment depends on domains, meshes, inflow, wall functions and convergence [3–5], with large-eddy simulation (LES) adding requirements for inflow turbulence and statistical sampling [6–11]. These operations recur across the wind directions and design options considered during planning. Pedestrian comfort assessment, for instance, combines direction-specific wind speeds at pedestrian height with comfort criteria [12]; the sensitivity of such assessments to urban morphology and to the number of simulated wind directions has also been examined [13].

Automation therefore has engineering value when it organises modelling, solution, post-processing and checks across cases while preserving links between those stages.

Comparability and recomputability must be established together in urban wind calculations. Inconsistent inlet profiles and wall functions alter downstream inflow development [14,15]; discretisation errors, validation conditions and observational uncertainty affect interpretation [16,17]. Parameter-sensitivity studies also show that computational settings influence ventilation predictions [18]. Multi-case comparisons are therefore required to hold non-target conditions fixed so that differences correspond to the factor under study. In the roughness-sensitive urban LES task of this benchmark, for example, only surface roughness changes across three cases, while inlet speed, wind direction, mesh and turbulence model remain fixed. The requirement applies to solver configurations, sampling locations, statistical windows and reported metrics.

A reported wind-speed ratio can be checked only if its computational conditions, raw outputs and processing are identifiable. Solver completion alone does not secure this chain. An omitted sampling time dimension, an artefact from an earlier run or a changed fixed condition can obstruct recomputation and comparison even after all runs finish. The computational delivery of urban wind tasks is therefore studied: agent-produced results must remain checkable against requirements, run records and raw data after solution. Here a result is termed usable when its reported metrics can be recomputed by an independent scorer from the delivered raw outputs, as indicated by a positive numerical score.

Large language models (LLMs) bring reasoning and tool use to multi-step tasks [19,20], including chemistry and mechanics [21–23], building energy modelling, assisted indoor airflow design and structural design [24–27]. SWE-bench and AgentBench have made code repair and interactive execution subjects of systematic evaluation [28,29]. Urban wind CFD adds two requirements to such automation: constraints must stay active during execution, and reports must be connected to the computations actually performed.

Existing CFD agents provide a starting point for these questions. MetaOpenFOAM, OpenFOAMGPT, Foam-Agent and ChatCFD configure, run and correct OpenFOAM cases from natural language [30–33], while CFDAgent simulates complex flows with an immersed-boundary solver [34]. CFDLLMBench evaluates knowledge, numerical reasoning and workflow execution, including execution and solution quality [35]. These studies are extended here by checking the conditions shared across cases in an urban wind task and whether raw samples support independent metric recomputation.

Three questions structure the study. First, does a completed simulation run yield a usable urban wind result, and how can this be checked? Second, how can constraint checks, evidence reading and submission feedback enter LLM execution? Third, how do delivery quality and call costs change with the same model name and budget? Benchmark and scoring design address the first question, WindPilot addresses the second, and task scores, cost statistics and repair traces address the third.

Three contributions are made by UrbanWind-AgentBench and WindPilot, an execution framework (harness), that is, the program around the LLM that determines what the model sees, which tools it calls, when checks occur and when execution stops. First, 10 engineering families organise 200 tasks and 367 cases, with public contracts fixing variable factors, unchanged conditions, numerical thresholds and deliverable specifications. Second, structural checks and numerical recomputation evaluate completeness, run provenance, metric recomputation and reference agreement, supported by independent references for the 30 evaluated tasks. Third, contract parsing, execution provenance, deterministic verification and bounded acceptance review are organised outside the LLM so that feedback identifies actual configurations and deliverable files. Quality, cost and repair processes can then be compared under the same model name.

The rest of this paper is organised as follows. Section 2 reviews related work, Section 3 describes benchmark construction, Section 4 presents WindPilot, and Section 5 specifies the experiments. Sections 6 and 7 present the results and discussion, respectively, and Section 8 concludes the paper.

## 2 Related work

### *2.1 Urban wind CFD and best practice*

Urban wind CFD already has systematic guidance for modelling and checking. The COST 732 best-practice summary and Architectural Institute of Japan (AIJ) guidelines address domains, meshes, boundary conditions and convergence checks [3,4]; AIJ collaborative projects provide comparison data for isolated buildings and building groups [36,37]. Blocken summarises ways to improve reliability in urban-physics CFD and discusses the applicability and cost of LES and Reynolds-averaged Navier–Stokes (RANS) simulations [5,8].

Different urban wind applications require different statistics from the same flow field. Pedestrian comfort [12,13,38,39], pollutant dispersion [40–42], natural ventilation [11,18] and microclimate [2,9], together with complex-structure wind assessment and field-measurement correction [10,43], can draw on different heights, windows and spatial regions. Task definitions must therefore retain both physical conditions and statistical conventions so that post-processed values can be traced back to their samples and settings.

The Parallelized Large-Eddy Simulation Model (PALM) supports building-resolving simulations, surface energy balance and multiscale nesting; its dynamic core has been compared with wind-tunnel experiments for a neutral urban boundary layer [44–47]. For LES, inflow-turbulence generation [6,7], cube-array flows [48,49] and dynamic subgrid-scale models [50] provide methodological foundations. Discretisation-error assessment, verification and validation, and records of experimental conditions and measurement uncertainty support interpretation at different levels [16,17,51]. These engineering requirements are translated here into configuration constraints, sampling definitions, evidence requirements and reference comparisons within tasks.

### *2.2 LLM agents and benchmarks*

Research on LLM agents has expanded from action generation to execution interfaces and feedback organisation. ReAct interleaves reasoning and actions [19], Toolformer studies tool use [20], Reflexion uses linguistic feedback to guide subsequent attempts [52], CodeAct uses executable code as the action space [53], and MetaGPT organises workflows through multi-agent roles [54]. Their components and applications have been reviewed systematically [55]. SWE-agent treats the agent–computer interface as a design object [56], showing that interfaces outside the model also affect task completion. Task contracts, raw-data tools and submission feedback are organised outside the LLM in the same spirit.

Success criteria follow from what a benchmark asks an agent to do. Repository repair in SWE-bench [28], interactive environments in AgentBench [29], general-assistant work in GAIA and computer operation in OSWorld [57,58] each have different end conditions. ScienceAgentBench, SciCode and MLE-bench cover scientific discovery, research coding and machine-learning engineering [59–61], while FEABench addresses multiphysics finite-element modelling [62] and EngiWorld evaluates agents on 1,301 tasks in professional engineering software, with domain verifiers checking the resulting artefacts [63]. Their acceptance evidence ranges from repository tests to system states and numerical references. Urban wind tasks add two relationships central to this benchmark: fixed conditions across cases and an explicit path from raw output to reported metrics.

Semantic review complements programmatic checks for open-ended deliverables. MT-Bench, G-Eval and related surveys examine LLM–human agreement and judgement biases [64–67]. In this work, LLM review is placed after programmatic verification to interpret evidence and conclusions in task context, whereas task scores are calculated by a separate scorer after execution.

Interpreting agent performance also requires quality, cost and reference access to be reported together. AI agents that matter identifies cost and overfitting concerns [68], BetterBench examines benchmark quality and reproducibility [69], and contamination research shows why exposure of evaluation materials matters [70–72]. Accordingly, execution-visible contracts and checking rules are separated from final-scoring references, traces and deliverables are preserved, and action steps and tokens are reported.

### *2.3 CFD agents*

Machine learning has been used for flow modelling, prediction and enhancement of numerical methods [73,74]. For CFD automation, MetaOpenFOAM organises case configuration and execution through multiple agents [30], OpenFOAMGPT introduces retrieval-augmented generation [31], and Wang et al. compare the pass/fail outcomes and API costs of different LLMs within OpenFOAMGPT [75]. OpenFOAMGPT 2.0 further investigates end-to-end automation [76]. Dong et al. fine-tune an LLM on paired natural-language requests and OpenFOAM configurations [77]; CFD-copilot generates configurations with a domain-adapted LLM, corrects them through multiple agents and connects post-processing tools through the model context protocol (MCP) [78].

Foam-Agent assigns planning, meshing, input-file writing, execution, review and visualisation to specialised components, with dependency-aware file generation [32]. ChatCFD combines a structured knowledge base, error localisation and iterative reasoning, evaluating execution success and physical fidelity [33]; CFDAgent uses a language-guided multi-agent workflow for complex flows [34]. CFDLLMBench uses CFDQuery, CFDCodeBench and FoamBench to assess domain knowledge, numerical and physical reasoning with code generation, and OpenFOAM workflow execution, respectively [35]. Reviews of CFD agents also discuss numerical verification, workflow provenance and system organisation [79]. The methodological distinction of the present work lies in constraints and checks specific to urban wind tasks.

Foam-Agent achieves 88.2% execution success on 110 basic FoamBench cases and 62.5% on 16 advanced cases, alongside flow-field-error comparisons [32]. ChatCFD reports 82.1% execution success and 68.1% overall physical fidelity for 205 cases from OpenFOAM tutorials and the OpenFOAM wiki and 110 perturbed cases; fidelity combines conditions, physical models and flow-feature checks, and separate literature cases test complex problems [33]. In the present evaluation, comparison conditions across cases within a task, raw-sample support for delivered metrics and independent recomputation after submission are examined. Table 1 locates these objects within the broader task coverage of prior studies.

**Table 1.**
Comparison of CFD agent evaluation studies

| Study | Evaluation scale | Task or case sources | Specific evaluation focus | Feedback and final evaluation |
|---|---|---|---|---|
| MetaOpenFOAM [30] | 8 tasks, each tested 10 times | OpenFOAM tutorials; separate parameter-modification test | Case configuration, execution and requirement satisfaction | Multi-agent correction; executability levels, requirement checks and pass@1 |
| Dong et al. [77] | 21 cases | Tutorials, variants and additional flow cases | Domain model adaptation and case solution | Initial execution, iterative correction and final solution |
| Foam-Agent [32] | 110 basic + 16 advanced | FoamBench | End-to-end OpenFOAM workflows | Review and correction; execution success and flow-field error |
| ChatCFD [33] | 205 tutorial + 110 perturbed cases; 2 additional literature cases | Tutorials, OpenFOAM wiki, perturbations and literature problems | Configuration, execution and physical fidelity | Iterative correction; execution success and checks of conditions, models and flow features |
| CFDLLMBench [35] | 90 question-answer, 24 coding and 110 + 16 workflow tasks | Domain knowledge, numerical code and FoamBench | Knowledge, reasoning and workflow capabilities | Accuracy, execution, numerical error and, for coding tasks, convergence |
| CFD-copilot [78] | Representative aerodynamic cases | Engineering benchmarks, including aerofoils | Domain-adapted configuration generation, correction and MCP-connected post-processing | Automatic correction and comparison of numerical results with references |
| UrbanWind-AgentBench | 200 constructed; 30 evaluated | Urban wind engineering tasks | Within-task comparison constraints and recomputable deliverables | Execution checks and review; structural scoring and numerical recomputation |

Note: Evaluation scales retain the task, case or repeated-test units reported in each study and do not support direct success-rate comparisons. pass@1 is the proportion of tasks solved in the first attempt. The 200 tasks describe benchmark construction; the experimental comparison uses 30 OpenFOAM tasks with independent references.

## 3 UrbanWind-AgentBench

### *3.1 Design objectives*

The task is the organising unit of UrbanWind-AgentBench (Fig. 1). Each contains instructions, a public contract, inputs and a scorer, and can comprise several separately configured and solved cases. Tasks define research questions and final evaluation; cases implement computations. Variable factors, fixed conditions and aggregate metrics connect cases within a task.

Three design objectives connect that structure to evaluation. The 10 engineering families provide workflow coverage. Public contracts make constraints checkable against configurations.

Raw evidence, processing records and scorers make reported results recomputable from their sources. Agents work with tools in isolated environments and submit common files plus task-specific results; post-execution scoring assigns 0–100 from those deliverables and references.

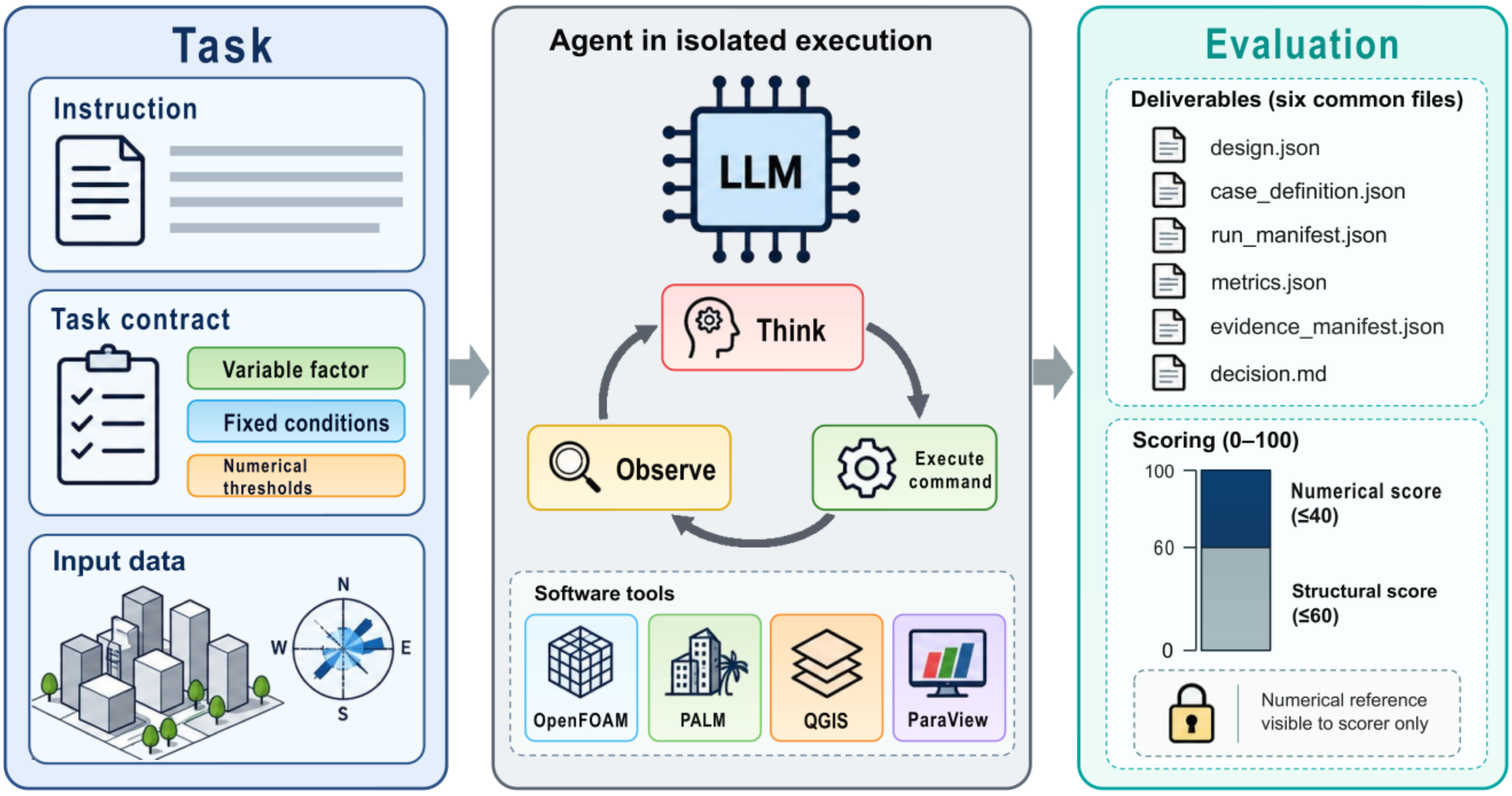

**Fig. 1.** Overview of UrbanWind-AgentBench

*3.2 Task families and difficulty*

Workflow coverage comes from 10 families of 20 tasks, giving 200 tasks in total (Table 2). Each engineering stage (materials, geometry, meshing, boundaries, solution and interpretation) has an explicit entry point, while outputs continue to inform later stages. Geometry and roughness guide meshes and inflows; solved fields support verification, mapping and design comparisons. The empirical evaluation in Section 5 uses the subset of tasks with established references.

Within each family, operational steps retain their engineering comparisons. Coarse, medium and fine meshes support grid convergence index (GCI) analysis; atmospheric boundary layer (ABL) tasks address log-law profiles, stability, synthetic turbulence and wind-direction transformations; RANS/LES tasks vary roughness, convection schemes or thermal forcing with other conditions fixed. Across families, inputs, settings, samples and metrics must stay linked; Table 2 lists the software and typical outputs that realise these relationships.

Task mode, domain difficulty and computational workload describe requirements, engineering complexity and resource demand separately. Modes comprise 53 configuration-and-verification tasks, 40 comparative simulations, 27 diagnostic repairs, 20 evidence audits, 20 physical-validation tasks and 20 post-processing-and-decision tasks; the remaining 20 involve design screening, optimisation and Pareto selection. Overall complexity labels basic, intermediate and advanced apply to 15, 44 and 141 tasks, respectively. Domain difficulty D1–D5 instead follows engineering-stage complexity: D1 denotes a single tool operation and D5 includes multi-case comparisons or engineering decisions. Counts are 15, 24, 20, 81 and 60.

By workload, 106 tasks use lightweight computation, 84 require CFD solution and 10 require large-scale solution. Together, the tasks contain 367 cases and 227 planned solves, each solve denoting one OpenFOAM or PALM solver run. There are 121 single-case tasks; the other 79 contain 2–12 cases. Each family involves at least 2 software packages (Fig. 2).

**Table 2.**

The 10 task families in UrbanWind-AgentBench (GIS, geographic information system)

| Task family | IDs | Engineering stage | Main software | Typical deliverables | Tasks |
|---|---|---|---|---|---|
| Evidence auditing and case definition | 001–020 | Extract geometry, inflow and sampling definitions | Python, QGIS | Parameter tables and sampling conventions | 20 |
| GIS and urban geometry | 021–040 | Process building footprints, heights and computational domains | QGIS, GDAL, Gmsh | Repaired geometry and roughness zones | 20 |
| Meshing and numerical discretisation | 041–060 | Block meshes, local refinement and mesh quality checks | OpenFOAM, Gmsh | Meshes and quality reports | 20 |
| ABL and boundary conditions | 061–080 | Log-law inflows, stability, wind-direction transformations and wall conditions | OpenFOAM, PALM | Inlet profiles and boundary configurations | 20 |
| OpenFOAM RANS/LES execution | 081–100 | RANS/LES solution and statistics for isolated buildings and building groups | OpenFOAM, ParaView | Flow statistics and comparison metrics | 20 |
| PALM urban microclimate | 101–120 | Urban surface energy balance and thermal-environment LES | PALM, NetCDF | Static drivers and energy closure | 20 |
| Fault diagnosis and repair | 121–140 | Locate and repair faulty cases | OpenFOAM, PALM | Repair records and conservation checks | 20 |
| Numerical verification and uncertainty | 141–160 | Grid convergence and reference-data comparison | SciPy, PyVista | Error and uncertainty metrics | 20 |
| Flow diagnosis and risk mapping | 161–180 | Map pedestrian wind comfort and risk areas | PyVista, QGIS | Wind-speed-ratio maps and exposure metrics | 20 |
| Ventilation and thermal-environment optimisation and decision-making | 181–200 | Design screening, Pareto selection and engineering decisions | OpenFOAM, PALM, pymoo | Design comparisons and decision notes | 20 |

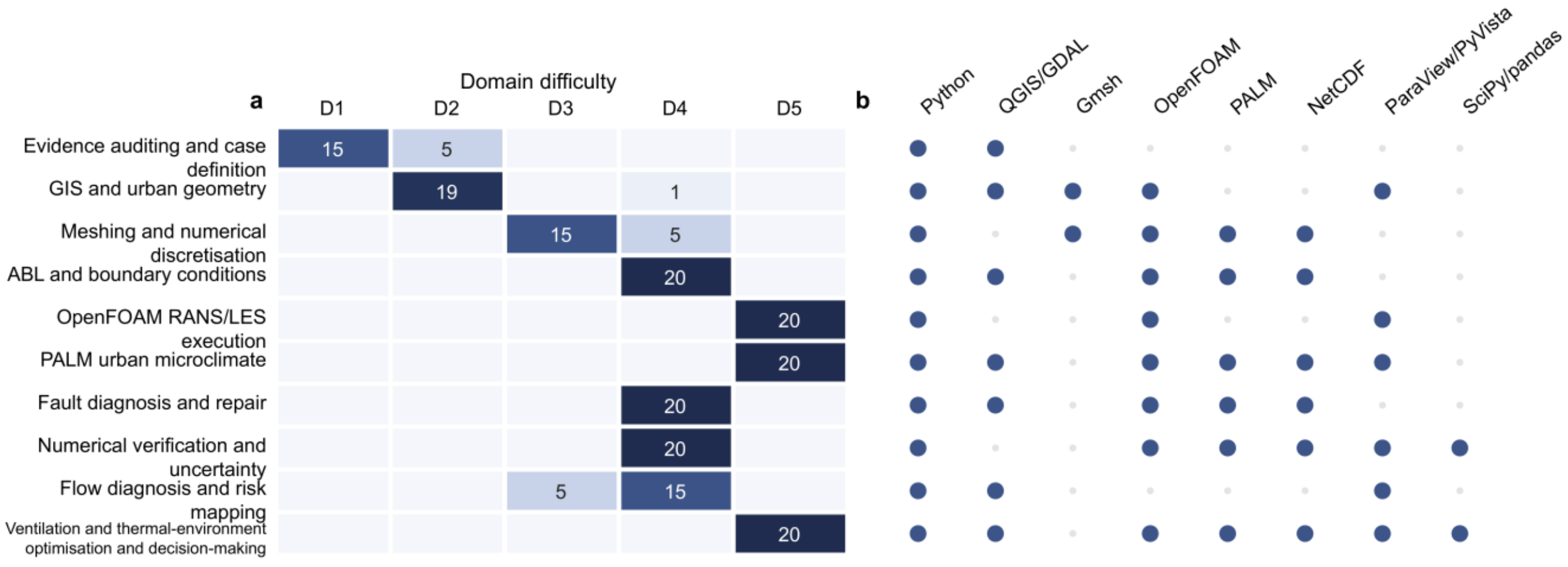

**Fig. 2.** Task distributions

(a) Task-family-by-domain-difficulty heatmap; cell values indicate task counts. (b) Software used by each task family; filled circles indicate use.

*3.3 Input data and initial states*

Inputs include script-generated, deterministic isolated buildings of width D and height 4D (denoted D×D×4D), 4×4 staggered cube arrays, procedural urban blocks, inlet profiles, OpenFOAM and PALM templates, synthetic observations and fault lists. Synthetic observations provide numerical comparisons; reference construction is described in Section 3.6. Public AIJ data are optional supplementary sources, and Fig. 3b groups geometry-source categories recorded in the tasks.

The initial state specifies how much workflow work remains. Engineering materials and models occur in 87 tasks, incomplete templates in 60, faulty cases in 27, from-scratch starts in 17 and runnable baselines in 9 (Fig. 3a). Materials and templates require definitions and settings to be completed, faulty cases require diagnosis, and runnable baselines permit checks or comparisons on existing computation. Seeded problems include mesh holes, scale-unit errors and incorrect subgrid parameters.

Geometry specifies the spatial problem independently of the workflow starting point. Procedural cities dominate 98 tasks, mainly supporting urban LES and thermal-forcing comparisons, while isolated buildings occur in 49 tasks for separation, recirculation and pressure statistics. Staggered cube arrays support canopy and above-canopy profile comparisons (Fig. 3b).

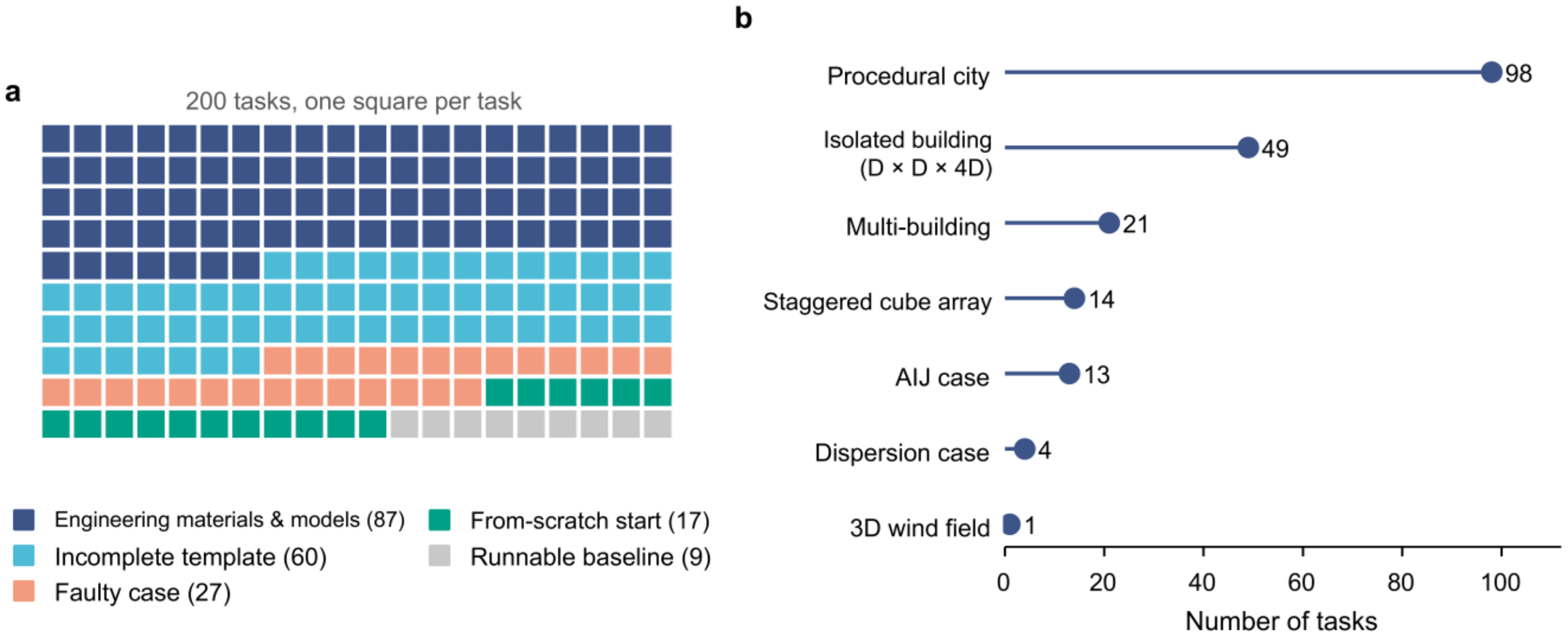

**Fig. 3.** Initial task states and geometry sources

(a) Waffle chart: each square represents 1 task, with colour indicating the initial state. (b) Task counts across 7 geometry-source categories.

*3.4 Task contracts*

Task contracts express engineering goals as four types of public requirements: variable factors and values, conditions to hold fixed, numerical thresholds and deliverable formats. Variable factors define comparisons, fixed conditions preserve comparability, thresholds specify public checks, and formats enable programmatic reading.

Six common files carry this contract into delivery: design.json and case_definition.json record tasks and settings; run_manifest.json records commands, software versions, logs and hashes; metrics.json and evidence_manifest.json link values to sources; and decision.md sets out conclusions, evidence and applicability. Task-specific files retain arrays or statistics beyond this common structure, including coordinate transformations for task-068 and simulation metrics for task-093.

The contract defines the object, formula, units and aggregation of the primary metric. For example, task-061 uses inlet-profile root-mean-square error weighted by probe-height intervals, in m/s. Probe heights, corresponding velocities and weights must therefore agree between reporting and recomputation. Multi-case tasks first calculate case-level values and then aggregate them according to the contract. All tasks also include evidence-provenance coverage and hard-threshold pass rates: traceable evidence items divided by required items, and passed hard thresholds divided by required thresholds, respectively.

Task-093 shows why contracts must span cases (Fig. 4). Its three urban LES cases compare aerodynamic roughness lengths $z_0 \in \{0.002\,\mathrm{m}, 0.02\,\mathrm{m}, 0.2\,\mathrm{m}\}$ with inlet speed, direction, mesh and turbulence model fixed. Checks must read those fixed fields from all actual cases, link raw probes to the corresponding roughness values and recompute sensitivity. Changing inlet speed in one case would introduce another factor.

Numerical thresholds also translate statistical decisions into specific data requirements. The transient spin-up and statistical-start task, task-087, uses wind-speed and turbulent-kinetic-energy statistics from 64 probes over 6 consecutive 60 s windows. It determines the earliest statistical start time under wind-speed drift ≤ 5% and turbulent-kinetic-energy drift ≤ 2%. Execution must therefore retain the corresponding time information so that final scoring uses the same statistical definitions.

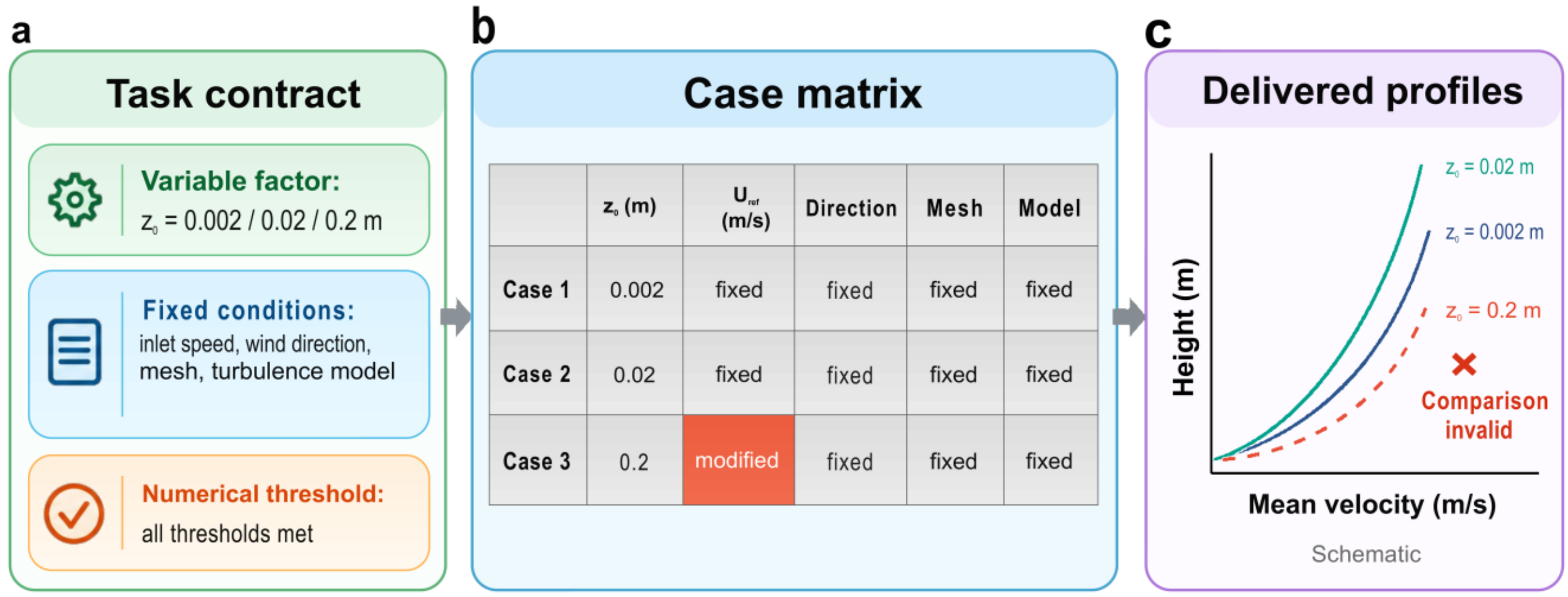

**Fig. 4.** Example task contract (task-093, roughness-sensitive urban LES). The curves on the right schematically illustrate the comparison; the central marker shows a contract mismatch caused by changing the fixed inlet speed.

### *3.5 Two-layer scoring*

Scoring separates delivery structure from task-specific numerical checking (Fig. 5; Table 3). The total is $S = S_s + S_n$, with structural score $0 \le S_s \le 60$ and numerical score $0 \le S_n \le 40$. The first layer checks readability and provenance organisation; the second checks numerical review and recomputation.

Structural scoring allocates 15 points to deliverable completeness, 10 to parseability, 15 to run provenance, 10 to raw evidence, 5 to metric traceability and 5 to the decision note. Completeness and parseability establish access for subsequent reading; provenance and raw evidence connect execution to artefacts; traceability and decision notes link reports to conclusions. For example, a run manifest without the solver name and version earns no provenance points, and omission of applicability conditions loses the corresponding decision-note credit.

Task-specific scorers assign numerical scores according to the engineering objective. Meshing tasks rerun the OpenFOAM mesh-quality utility checkMesh and earn 40 points for an acceptable mesh with complete boundaries. Comparative tasks check raw evidence, reported-array recomputation, primary-metric recomputation and reference agreement. For task-062 these receive 12, 8, 5 and 15 points. For task-093, result format and thresholds, metric recomputation and hidden-array agreement receive 8, 12 and 20 points, respectively (Table 3).

The two layers use raw evidence at different depths. Structural checks establish that outputs exist and reports point to them; numerical checks read those outputs and perform the specified calculations or comparisons. In task-093, raw probe data from three cases produce a recomputed roughness-sensitivity slope for reference comparison.

Scorers execute after the agent stops, and reference arrays and matching tolerances are used only in final scoring. A total above 60 therefore implies a positive numerical score.

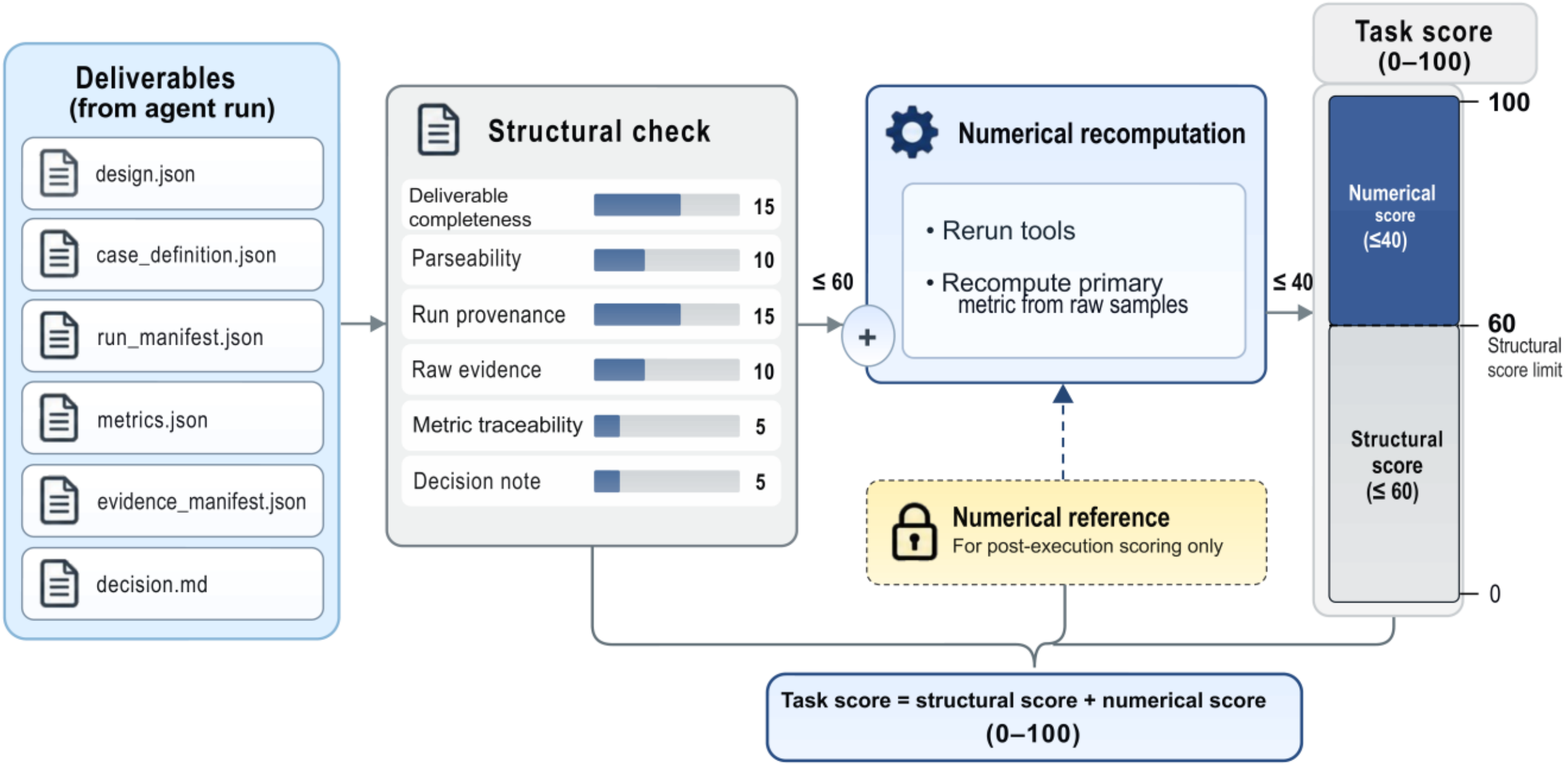

**Fig. 5.** Two-layer scoring. The structural and numerical scores are capped at 60 and 40 points, respectively, and sum to the total task score.

**Table 3.**

Scoring items and point allocations

| Layer | Scoring item | Points | Checks |
|---|---|---|---|
| Structural | Deliverable completeness | 15 | Presence of the 6 common files and task-specific files |
| Structural | Parseability | 10 | Programmatic readability of JSON and tabular files |
| Structural | Run provenance | 15 | Correspondence between commands, logs and artefacts in the run manifest |
| Structural | Raw evidence | 10 | Provision of raw solver outputs |
| Structural | Metric traceability | 5 | Traceability of reported metrics to raw files |
| Structural | Decision note | 5 | Presence of conclusions, supporting evidence and applicability conditions |

| Layer | Scoring item | Points | Checks |
|---|---|---|---|
| Numerical (task-062) | Raw profile evidence / reported-array recomputation / primary-metric recomputation / numerical-reference agreement | 12 / 8 / 5 / 15 | Profile-law checks, array agreement with raw samples, metric recomputability and errors relative to the reference |
| Numerical (task-093) | Result format and thresholds / metric recomputation / hidden-array agreement | 8 / 12 / 20 | Run-threshold compliance, primary-metric recomputability and probe-array error relative to the reference |
| Numerical (task-041) | Mesh recheck | 40 | Rerun checkMesh; award points for an acceptable mesh with complete boundaries |

Note: Structural scoring is identical across tasks. Numerical scoring items are defined by each task-specific scorer; 3 representative tasks are shown here.

### *3.6 Numerical references*

Reference construction used OpenFOAM v2412 solution and meshing tools, producing numerical references for 29 tasks and a mesh-tool reference for 1 task, across 56 cases and 53 solves. The tool reference uses checkMesh to recheck mesh status. Numerical references store probe coordinates, time windows and case-level primary metrics as arrays, enabling comparisons of equivalent computational objects. These references support task evaluation and are managed separately from public thresholds available during execution.

Task instructions, contracts and scorers underwent initial domain review, independent in-depth review, local data checks, numerical-reference checks and independent re-review. These stages checked contract consistency, input completeness and scores for reference solutions.

Configuration checks read boundary types, parameters and sampling intervals from specific OpenFOAM dictionary locations. The parser reads explicit dictionaries and reports unparseability when include directives or embedded code remain to be expanded, without executing that code. Structural checking tools are available during execution, whereas reference comparisons occur afterwards.

## 4 WindPilot execution framework

### *4.1 Design and overall architecture*

WindPilot keeps contracts, computations and submission feedback within the same execution loop. Contracts specify relationships to preserve, provenance associates runs with artefacts, deterministic verification reads files, and acceptance review checks evidence against conclusions.

Five modules implement this loop around LLM thinking, command execution and observation: isolated execution, contract parsing, execution provenance and delivery, deterministic verification, and bounded acceptance review (Fig. 6). Initial parsing supplies constraints and a delivery checklist. Commands create settings, logs and outputs linked by run records. Completion declarations initiate organisation and verification, with LLM review following a programmatic pass. Any repair request returns to execution, and the revised delivery is checked again.

Feedback names files, fields and difference locations so that it can direct further reading, edits or calculation. Final deliverables are saved for evaluation by a separate scorer.

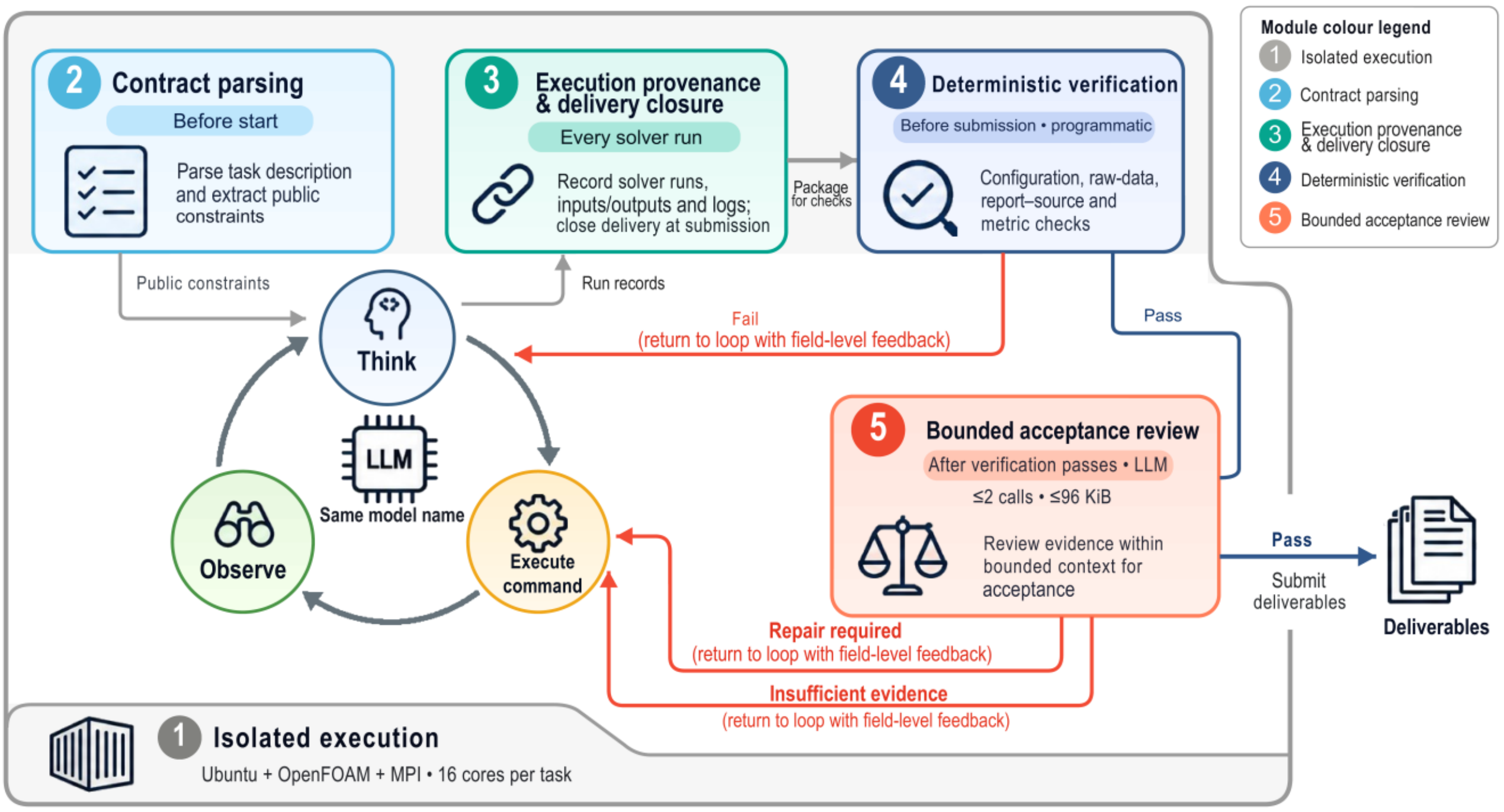

**Fig. 6.** WindPilot architecture

Public constraints enter through contract parsing, and run records link solver execution to artefacts. A completion declaration starts delivery checks; their errors and review issues return to the loop. The illustrated submission path follows checking, whereas budget-exhaustion termination and scoring are defined in Section 5.4.

**Table 4.**

Inputs and outputs of WindPilot modules

| Module | Objects processed | Key operations | Outputs and purpose |
|---|---|---|---|
| Isolated execution environment | Task workspace and CPU resources | Separate environments, core allocation and process management | Define the solver's execution boundaries |
| Task-contract parsing | Public constraints and deliverable specifications | Compile checklists and inspect actual fixed conditions | Keep constraints and interface requirements available |
| Execution provenance and delivery loop | Commands, inputs and artefacts | Record runs; organise IDs and hashes | Link deliverables to actual execution |
| Deterministic verification | Configurations, raw outputs and reports | Check dimensions, time axes, hashes and metrics | Return errors with identifiable locations |
| Bounded acceptance review | Task and deliverable evidence view | Assess whether evidence supports conclusions | Recommend submission or repair |

## *4.2 Contract parsing and deliverable specifications*

An acceptance checklist makes contract requirements persist during execution. Parsing gathers variable factors, fixed conditions, public thresholds and file formats into context available to the LLM as it reads outputs or revises settings. Runtime configuration checks then compare the specified fixed conditions with the fields actually used in solver files.

Contract prompts state the conditions that should remain consistent, and configuration checks locate mismatches by file and field. In a multi-case comparison, every participating case must satisfy each fixed condition, with only the contract-specified factors taking their allowed values.

For example, the convection-scheme sensitivity task, task-085, permits only the velocity-convection discretisation scheme to differ between two cases. Mesh, time step, subgrid-scale model, inlet conditions, solver tolerances and statistical window must remain identical. Contract checking separates these six fixed-condition categories from the permitted scheme change.

Deliverable specifications further define the root task ID, nested case containers, diagnostic fields, evidence roles, product types and array shapes. Runtime thresholds come from public contracts, while reference arrays and matching tolerances are reserved for post-execution scoring.

*4.3 Execution provenance and deliverable organisation*

Execution provenance links reported results to the solver runs and post-processing that generated them. Each command records input files, the command, working directory, software version, standard output and error, exit code, elapsed time and artefact hashes; these records form the run manifest. Commands and logs describe execution, files and hashes identify inputs and artefacts, and evidence references in the report connect metrics to these records.

When a configuration changes and the solver is rerun, final deliverables must correspond to the updated run. The manifest specifies which solve and processing results support the final metrics. Section 6.5 illustrates rebinding the final run record, post-processing script, statistical window and output hashes in task-094.

Raw-data tools provide repeatable computational access for these links. The framework supports probe and field-data reading, statistics over actual time windows, population-variance and turbulent kinetic energy (TKE) calculation, normalised mean absolute error (NMAE) under explicit masks, and pointwise dispersion across cases.

Log and OpenFOAM-dictionary summaries connect these calculations to solver settings and execution. Missing data or invalid formats produce an unavailable-calculation result with the problem's location and type.

Delivery organisation reconciles the files that carry provenance. It supplies task and case IDs, normalises count-field types and refreshes hashes affected by those edits, using existing files and public inputs. Preserved pre-edit bytes and modification records keep this organisation distinguishable from later numerical work.

Hash updates require a consistency check against pre-edit bytes. A hash is refreshed after organisation only if it matched the file before modification. Already-stale references therefore remain visible in verification feedback. Configuration changes, array re-export and solver reruns are left to subsequent execution actions.

*4.4 Deterministic verification and repair feedback*

Deterministic verification checks four relationships between contracts and deliverables: configuration compliance, raw-data completeness, report–source correspondence, and metric compliance with public calculation definitions and thresholds. Configuration checks read sampling objects and boundary names from the actual controlDict, the OpenFOAM run-control dictionary. Data checks inspect probe components, array dimensions, time axes and statistical windows. Provenance checks compare hashes and reported arrays. Metric checks recompute available primary metrics and compare reported values with public thresholds.

Returned issues identify what needs attention, so that record completion, array export or configuration changes can be chosen.

Array verification preserves time, spatial-point and component dimensions, compares reports with raw samples element by element, and returns the first difference location. Configuration verification reads end times, write intervals and sampling settings and checks boundary types and referenced profile files. Inlet-profile errors are recomputed with layer-thickness weights, and LES statistics use contract-defined windows. In task-068, the mismatch between a 52×3 velocity array and 20×52×3 raw samples directly identifies the missing time dimension; Section 6.5 describes its repair.

Checks are triggered by agent requests, completion declarations, entry to acceptance review and prescribed budget-reservation stages. Feedback identifies file, field path and first mismatch, grouping equivalent elementwise errors. Programmatically decidable failures return for repair without calling review, and revised files are checked again.

Check states are pass, fail and unknown. Missing public definitions or semantics that cannot be determined unambiguously are recorded as unknown. Each verification trigger, check item and issue is written to the execution trace, supporting the event-, task- and error-category statistics in Section 6.6.

*4.5 Bounded acceptance review and submission control*

After deterministic verification passes, bounded acceptance review examines whether supplied evidence supports task conclusions. It reads the contract, run manifest, metrics and summaries of raw logs, placing local programmatic findings in task context. Review addresses correspondence between actual boundaries and requirements, whether run records support current metrics, and whether the decision note cites sufficient evidence.

The review has three outcomes. Ready to submit indicates support from the reviewed evidence; repair required identifies located defects; insufficient evidence identifies missing support. The latter two return the task to execution.

A length-limited evidence view structures review inputs. If capacity is exceeded, task contracts, deterministic-verification results and run records take priority, and omitted material is marked unreviewed. The review prompt requires supplied evidence with consistent hashes and file- and field-specific issues. Table 6 lists evidence-view capacity and call limits.

A repair restarts deterministic verification on the new files, then proceeds to the next available review after a pass. Reviews and repairs consume the shared execution budget; reaching an action or time limit preserves the current delivery for scoring.

*4.6 Isolated execution and budget management*

Each task runs in a separate Ubuntu environment with OpenFOAM, the Message Passing Interface (MPI) and common Python scientific libraries preinstalled; networking and software installation through apt and pip are allowed. The scheduler assigns non-overlapping physical CPU cores and manages solver processes through processor affinity and process-ownership locks. Platform parameters appear in Section 5.4.

Budget management jointly constrains execution, checking and repair. A runtime guard limits foreground-command duration, reminds the agent of remaining actions and time each round, and recognises the solver-startup trapFpe activation message as normal output. Delivery checks start after a preset action threshold to reserve resources for feedback. Tool CPU time, LLM tokens and wall-clock time are recorded with execution; action thresholds and time parameters are given in Table 6.

# 5 Experimental design and evaluation methods

## *5.1 Evaluation objectives and task set*

The evaluation follows three levels of evidence. Overall scores and distributions compare full WindPilot with the base agent, in which the same model runs only a think–execute–observe command loop without contract parsing, verification or review. Engineering-type breakdowns examine where quality and cost change. Staged configurations and task traces then show how contracts, raw evidence and submission feedback enter execution.

The 30 tasks comprise 1 meshing/discretisation, 13 ABL/boundary and 16 OpenFOAM RANS/LES tasks, with 56 cases. Modes comprise 12 configuration/verification, 2 diagnostic repair and 16 comparative simulation tasks; 15 contain 2–3 cases. Domain difficulty is D3 for the meshing task, D4 for ABL tasks and D5 for RANS/LES tasks.

Coverage includes task-041; ABL tasks task-061–064, 068–069, 071–072, 074–076 and 078–079; and RANS/LES tasks task-082–085, 087–089 and 091–099. Engineering content spans meshes, neutral and stability-dependent inflows, direction transforms, surface heat flux and temperature, roughness/convection sensitivity, thermal forcing, passive scalars and decomposition reproducibility.

## *5.2 Comparison conditions*

Both the base agent and WindPilot call DeepSeek-V4.1-Flash through the official application programming interface (API). Acceptance review uses the same model name, and sampling parameters retain API defaults [80,81]. Both conditions use identical task packages, physical conditions, public thresholds and final scoring rules, together with the same action caps and time-budget rules. The experiment therefore compares complete execution configurations with the same model name and resource limits.

The base agent organises work from task materials through an LLM command-interaction loop. WindPilot surrounds that loop with contracts, provenance, delivery organisation, deterministic checks and acceptance review, plus raw-data and statistical tools.

The structural tools in Section 3.6 check deliverable structure. WindPilot additionally reads actual configurations and raw outputs at designated events, returns findings to execution, and reviews evidence after programmatic verification passes. A completion declaration ends base-agent execution, whereas WindPilot first checks and returns feedback. This termination difference is part of the execution configuration being compared.

## *5.3 Staged configuration comparison*

Four staged configurations describe the added function bundles (Table 5): base agent; isolated execution and contract prompts; verification/review and runtime guard; and full WindPilot. Environment and prompting change first, checking feedback, review and runtime management follow, and delivery closure, evidence consistency and primary metrics are added last. Section 6.4 reports the scores for this sequence.

Stage coverage is 30, 28, 30 and 30 scored tasks. Each available-task mean is reported, and additionally the isolated-execution-and-contract-prompt stage is compared with the base agent on 28 common tasks.

**Table 5.**

Staged execution configurations and evaluation coverage

| Configuration | Functions added relative to the preceding stage | Tasks with score records |
|---|---|---|
| Base agent | LLM command-interaction loop | 30 |
| Isolated execution and contract prompts | Isolated execution environment and task-contract prompts | 28 |
| Verification, review and runtime guard | Acceptance checklist, deterministic verification, bounded acceptance review and runtime guard | 30 |
| Full WindPilot | Delivery loop, evidence-consistency verification and primary-metric checks | 30 |

### *5.4 Runtime environment and budget settings*

The platform is an AMD EPYC 7B12 server with 128 physical cores; each task receives 16 cores, using OpenFOAM v2412 and MPICH. Workspaces are initialised from environment snapshots and receive task inputs, templates and public contracts. Scorers and references are uploaded after execution for scoring. Initialisation, execution and scoring follow this order; Table 6 summarises platform and budget parameters.

A task can execute at most 150 steps and retains its latest 15 interaction rounds, with a time budget of 5 h of operations plus twice its historical solve time. WindPilot initiates delivery checks at step 90 to leave actions for repair; actual check positions are logged. Completion ends the base condition, whereas WindPilot follows Sections 4.4–4.5 before submission.

When actions or time run out, current deliverables are retained for scoring. Each task permits at most 2 acceptance reviews and reserves no more than 600 s for delivery. Checking, review and subsequent repairs all use the overall execution budget. Scoring has a separate 900 s allowance outside the agent execution budget.

Each task runs once under each condition, with commands, standard outputs, exit codes, interaction traces and final files preserved. Runs incomplete because of platform or API failures are missing rather than scored zero; paired statistics use tasks with scores in both conditions. Each LLM call times out after 600 s and permits at most 3 retries on failure. Review tokens and time are included in execution statistics.

**Table 6.**

Runtime environment and budget parameters

| Category | Parameter | Setting |
|---|---|---|
| Computing environment | Processor and per-task allocation | AMD EPYC 7B12; 128 physical cores in total; 16 cores per task |
| Computing environment | Solver and parallel library | OpenFOAM v2412; MPICH |
| Model calls | Execution and acceptance-review model | DeepSeek-V4.1-Flash; official API; default sampling parameters |
| Model calls | Per-call timeout and retries | 600 s timeout; up to 3 retries on failure |
| Execution budget | Action limit and history | At most 150 steps; retain the most recent 15 interaction rounds |

| Category | Parameter | Setting |
|---|---|---|
| Execution budget | Total time budget | 5 h of operational time + twice the historical solver runtime |
| Runtime guard | Foreground command limit | 300 s |
| Delivery and review | Check initiation and reserved time | Start at step 90; reserve at most 600 s for delivery |
| Delivery and review | Acceptance-review limits | At most 2 calls per task; evidence view at most 96 KiB |
| Run count | Per task per condition | 1 |
| Independent scoring | Scoring-stage limit | 900 s; excluded from the agent execution budget |

*5.5 Evaluation metrics and statistical analysis*

Task totals from Section 3.5 quantify delivery quality through means, medians, family distributions and counts of tasks scoring above 60, at least 80 and 100. Exceeding 60 establishes numerical credit; component scores identify which recomputation and reference checks produced it.

The task is the unit of analysis: the 56 cases are aggregated into task-level metrics under each contract and are not counted as independent observations.

Execution costs include per-task action steps, tokens and the number of tasks exhausting 150 steps, with paired directions of change. Action steps count executed commands, and tokens sum API-reported usage for execution and acceptance review. Tool CPU and wall-clock times are preserved in run records for checking specific computational processes.

The score comparison averages within-task differences, subtracting the base score from WindPilot's score. A task-level paired bootstrap resamples these differences with replacement, retaining the original set size, for 10 000 replicates. The 2.5% and 97.5% positions of sorted replicate means define the 95% interval. Overall and ABL calculations are performed separately with seed 20261024.

Process analysis distinguishes verification events, tasks and deduplicated errors. Events record triggers, action positions and detected problems. Identical error strings repeated within a task count once; errors absent from that task's final verification are defined as repaired. Categories include missing or unreadable files, report–raw-data inconsistency, stale hashes, incomplete decision notes, reported values exceeding public thresholds and invalid metrics. In case analysis, feedback is linked to subsequent commands and deliverable changes; review outcomes are aggregated by call.

## 6 Results

*6.1 Completed runs versus usable results*

A completed run rarely yielded a usable result under the base agent: at least the structural maximum of 60 points was reached on 28 of 30 tasks, yet numerical credit was obtained on only 6. Overall delivery scores on the 30 paired tasks were improved by WindPilot (Table 7; Fig. 7). Means were 72.5 versus 62.5, a 10.0-point difference with 95% interval [3.63, 16.67]. At least 80 points were obtained on 10 rather than 3 tasks, and 3 rather than 0 reached full scores.

Full scores came from task-041 (meshing), task-061 (neutral log-law inlet) and task-064 (Monin–Obukhov stability inlet). Their components are given in Section 6.2.

**Table 7.**
Main results (30 tasks)

| Metric | Base agent | WindPilot |
|---|---|---|
| Mean task score | 62.5 | 72.5 |
| Tasks with total scores >60 | 6 | 17 |
| High-scoring tasks (≥80) | 3 | 10 |
| Full-score tasks (100) | 0 | 3 |
| Mean action steps | 130.4 | 104.9 |
| Tasks exhausting 150 steps | 8 | 2 |

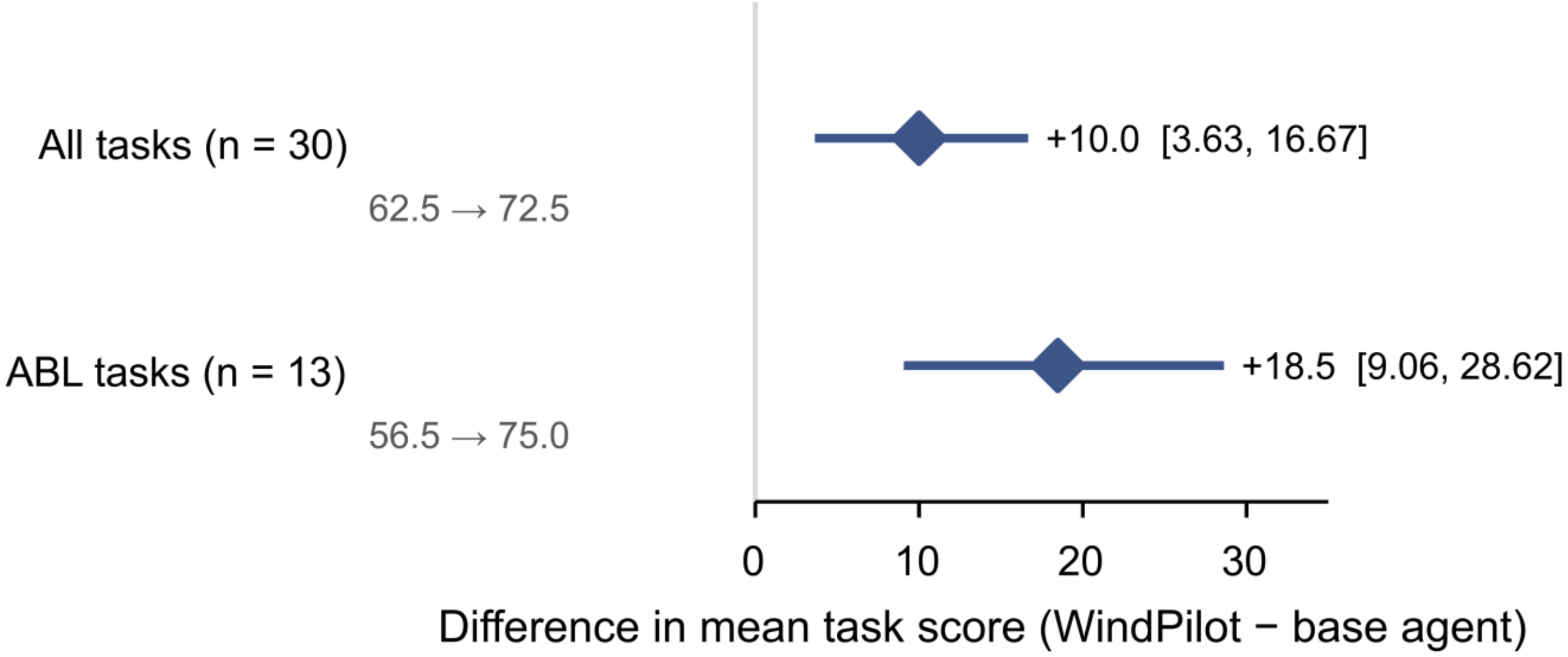

**Fig. 7.** Paired differences in mean task scores
Forest plot: diamonds indicate the mean task-score difference between WindPilot and the base agent; horizontal lines show 95% bootstrap confidence intervals (10 000 resamples). Mean scores under both conditions appear on the left.

In the score distributions (Fig. 8), the base agent has 22 tasks at exactly 60, 6 above 60 and 2 below 60, compared with 12, 17 and 1 for WindPilot. The median rises from 60.0 to 69.1 and the minimum from 30 to 55. At 60, the cumulative proportions are 0.800 and 0.433; WindPilot's curve is no higher throughout the score range.

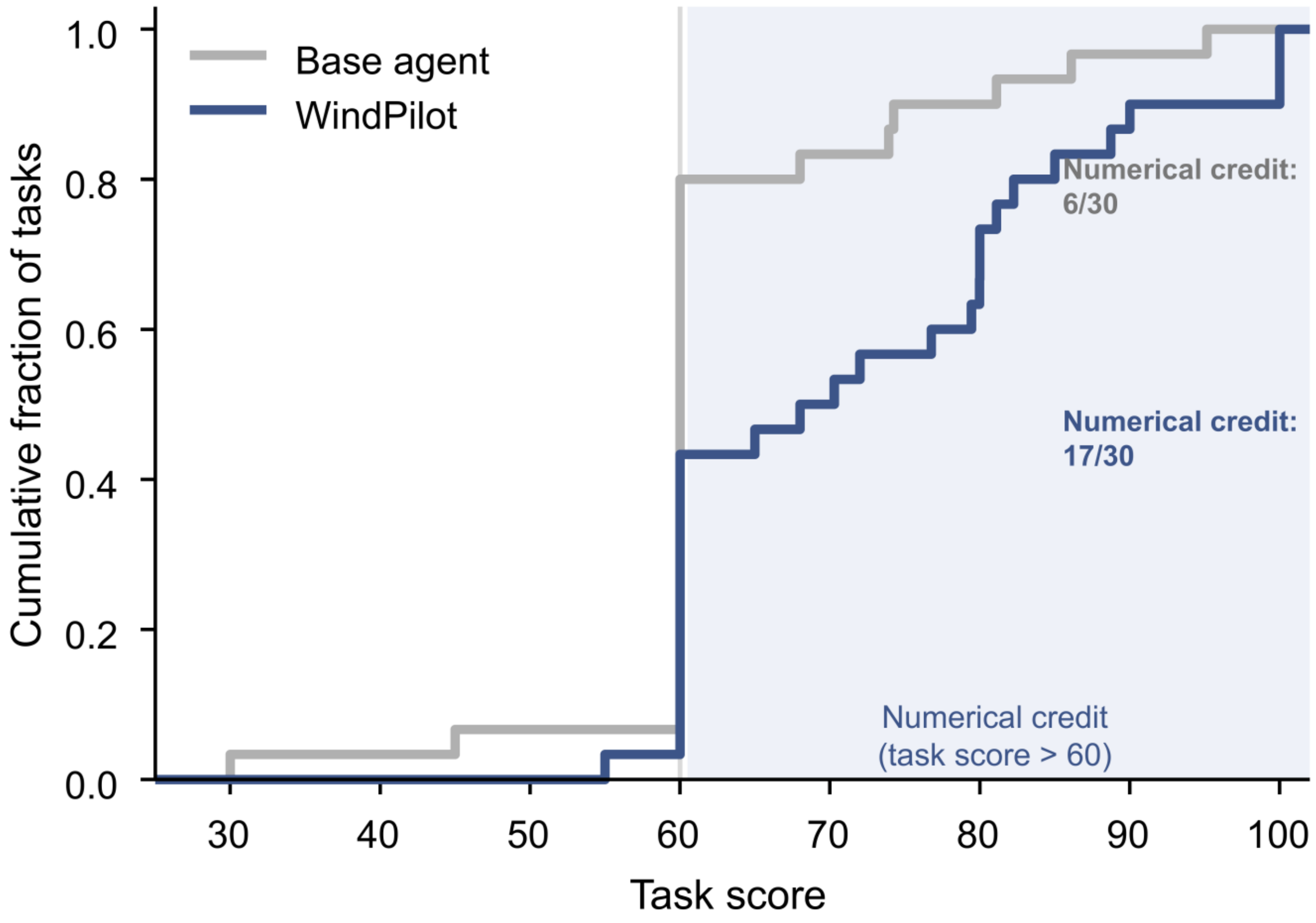

**Fig. 8.** Empirical cumulative distribution of task scores
The vertical axis is the proportion of tasks scoring at or below the horizontal-axis value; shading marks total scores >60, which guarantees a positive numerical score under the scoring rule.

*6.2 Gains across task types*

Inlet and boundary tasks accounted for most improvement. ABL means across 13 tasks rose from 56.5 to 75.0, a paired gain of 18.5 points (95% interval [9.06, 28.62]); meshing task task-041 rose from 60 to 100. RANS/LES means across 16 tasks changed from 67.4 to 68.6, calculated by removing ABL and meshing scores from overall totals. Summed ABL gains made up about 80.0% of the overall increase.

ABL improvement was distributed across tasks (Fig. 9): 9 improved, 4 tied and none declined. The group median rose from 60 to 72, the median per-task gain was 15, and >60 counts rose from 0 to 8. Examples include task-068 (30→90), task-064 (60→100) and surface-temperature task-075 (60→80).

Component scores explain the content behind representative high scores. Task-061 earned 15, 10, 5 and 10 points for raw neutral-ABL evidence, reported-array recomputation, primary-metric recomputation and reference agreement, respectively; the base agent earned zero on each.

In task-068, dimension, evidence-reference and run-record corrections accompanied the score gain; the repair sequence is given in Section 6.5.

In the RANS/LES group, tasks above 60 increased from 6 to 8 and the median from 60 to 64. These tasks involve longer solves and cross-case statistics, and their mean improvement was smaller than the ABL group's. Task-041's 40-point gain came from checkMesh rechecking. These group differences are discussed in Section 7.3.

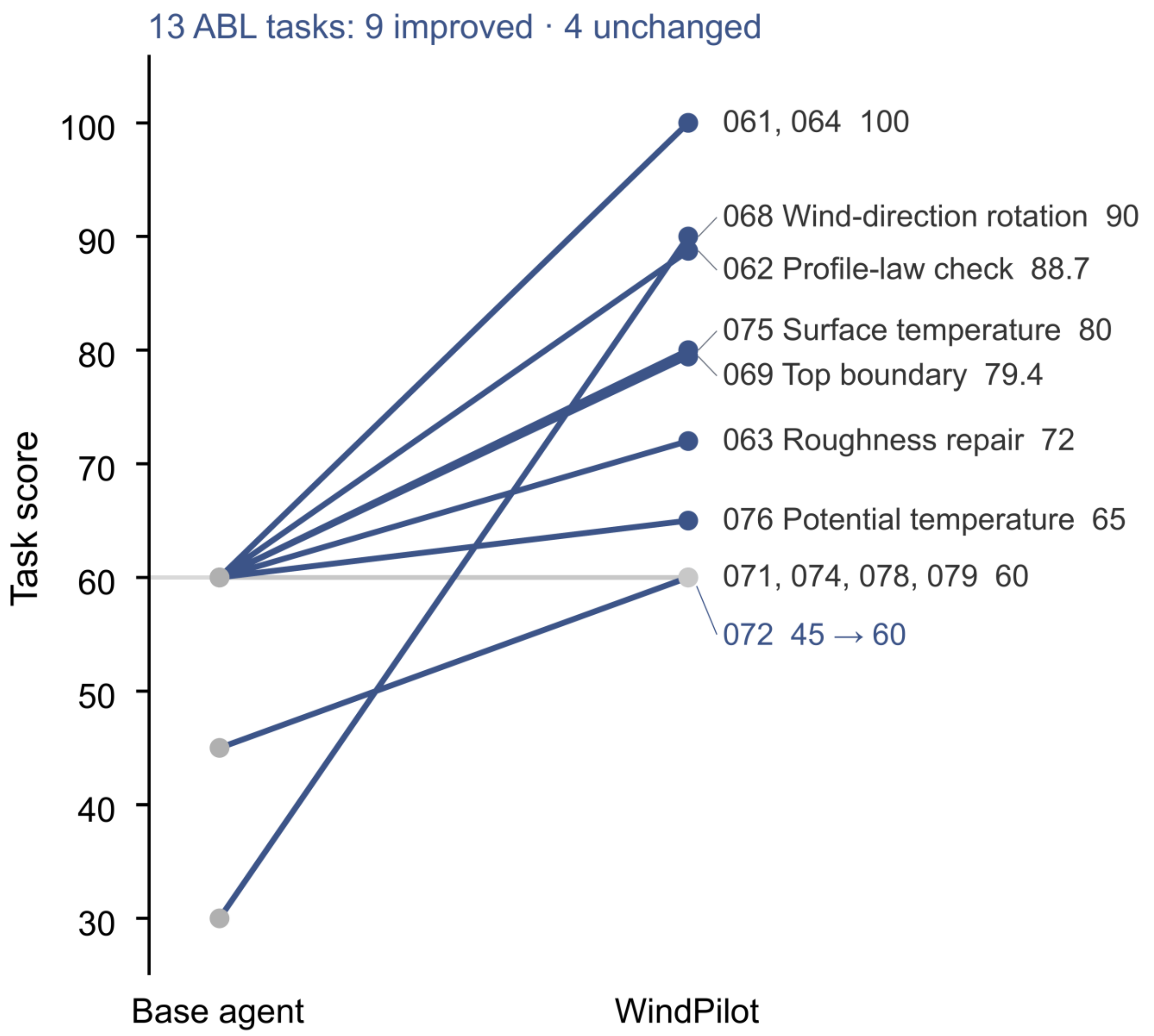

**Fig. 9.** Scores for individual ABL tasks

Slope plot: each line joins the scores for one task under the two conditions. Blue indicates improvement and grey indicates no change.

*6.3 Action counts and model-call costs*

Fewer interactions accompanied the score improvement. Mean actions decreased from 130.4 to 104.9 (19.5%) and the median from 135.5 to 103.5; cap-exhausting tasks decreased from 8 to 2 (Fig. 10). Action steps decreased for 24 tasks, were unchanged for 1 and increased for 5. Seven of the 8 previously capped tasks ended below 150 with WindPilot.

Both main groups reduced actions, with larger reductions for ABL. Its mean fell from 133.5 to 99.1 (25.8%); 12 of 13 tasks decreased and capped tasks fell from 5 to 0. RANS/LES fell from 129.1 to 112.7 (12.7%), with 11 decreases, 1 tie and 4 increases; capped tasks fell from 3 to 2. ABL thus combined larger score gains with larger action savings.

Model consumption likewise fell, including review calls. Mean tokens decreased from $6.29\times10^6$ to $5.48\times10^6$ (12.9%), with 20 taskwise decreases and 10 increases. ABL and RANS/LES mean reductions were 17.0% and 8.7%.

Interaction savings do not fully describe solver work. The number of cases launched in a command, solve length and reruns also determine CFD computation. Task-094 reran three cases after feedback (Section 6.5), illustrating how improved delivery can require more numerical work. Separate reporting of actions, tokens and solver processes preserves this distinction.

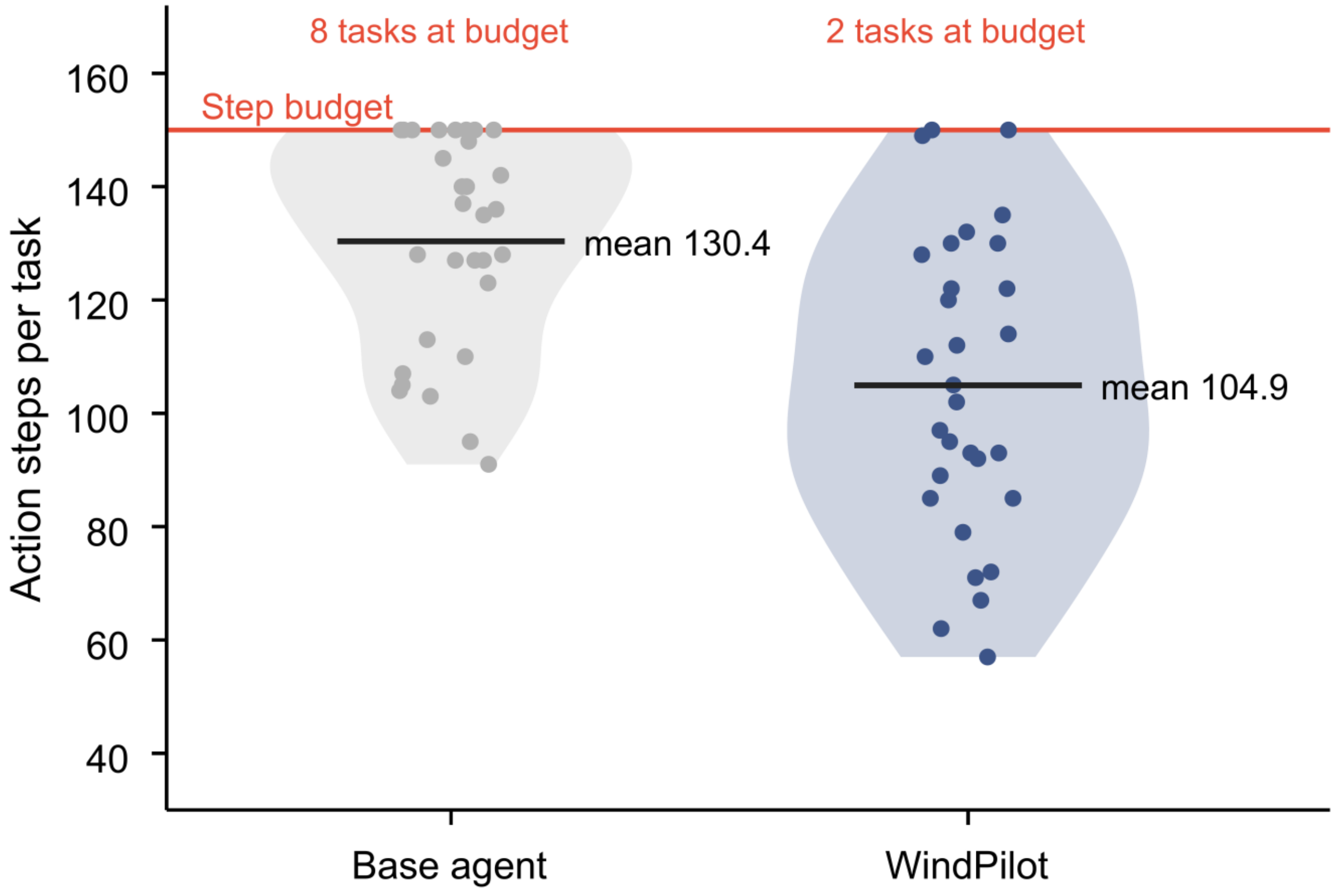

**Fig. 10.** Distribution of action counts

Violin plots with points, each representing one task. Black horizontal lines indicate means, the red line marks the 150-step limit, and counts of tasks exhausting the limit appear above.

*6.4 Task scores under different execution configurations*

Staged means rose mainly after checking and review entered execution (Fig. 11). Available-task means were 62.5, 62.3, 70.5 and 72.5 for the base, isolated-execution/contract-prompt, verification/review/runtime-guard and full configurations, based on 30, 28, 30 and 30 tasks. On their 28 common tasks, the first two scored 62.6 and 62.3.

Verification/review with runtime guarding reached 70.5, and full WindPilot reached 72.5 after adding delivery closure, evidence-consistency checks and primary-metric checks. These additions address report–output relationships through IDs, source hashes, dimensions and metric-calculation access points.

Adjacent differences in Fig. 11 are arithmetic differences between the means of function bundles.

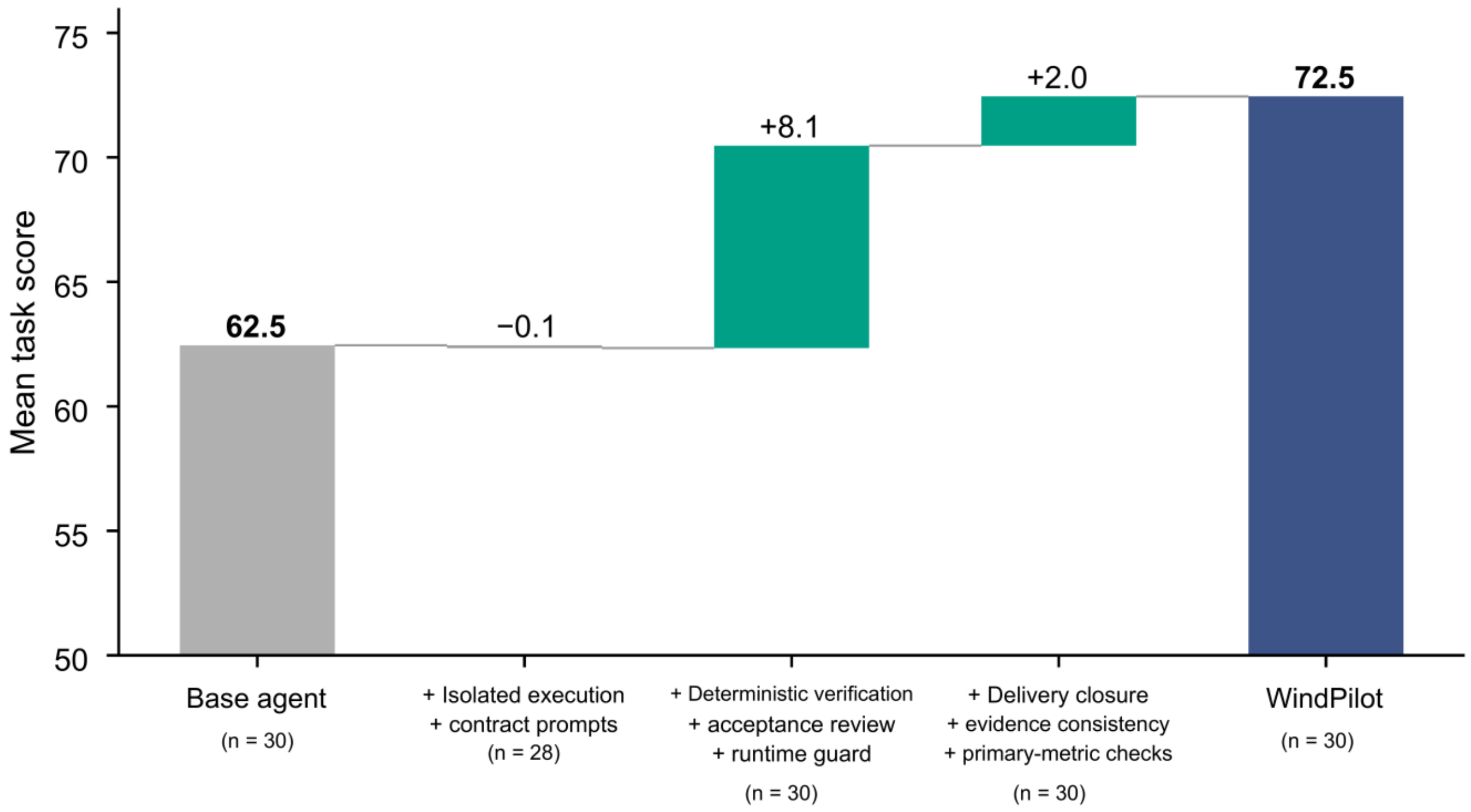

**Fig. 11.** Mean task scores under different execution configurations

Waterfall chart of observed means and adjacent arithmetic differences. Sample sizes are 30/28/30/30 and means 62.5/62.3/70.5/72.5; differences use unrounded means. Intermediate bars denote jointly added functions, and the last bar restates the full-framework mean.

### *6.5 Error localisation and deliverable repair in representative tasks*

Task-094 tests whether a three-case thermally forced urban LES comparison can be recomputed. It compares vertical exchange under neutral and two thermal-forcing scenarios, requiring conditions, raw velocity probes and processing records for each primary metric. The base agent solved all cases and submitted at step 137, but omitted the first case's raw velocity probes from the evidence manifest. The primary metric could not be recomputed, and the task scored 60.

WindPilot's trace links feedback to repair (Fig. 12). A reserved-budget check ran at step 91, followed at step 92 by a completion declaration that passed deterministic checks. Review then found a ground heat-flux boundary inconsistent with the contract, continuity-error reports inconsistent with logs, and incomplete solver/raw-output provenance.

The agent adjusted near-wall thermal boundaries and reran all three cases, declaring completion again at step 115. The second review found that the manifest did not identify actual solver records and that registered post-processing could not reproduce delivered metrics. The agent then rebound final run records and supplied the post-processing script, statistical window and output hashes before submission at step 130. Raw evidence, reported-array recomputation and primary-metric recomputation earned 12, 8 and 5 points, respectively, producing a final score of 85.

The resulting 25-point gain corresponds to three repaired numerical components: raw evidence, arrays and primary metrics. Boundary changes affected actual computation, while record rebinding and added scripts connected final metrics to updated solves.

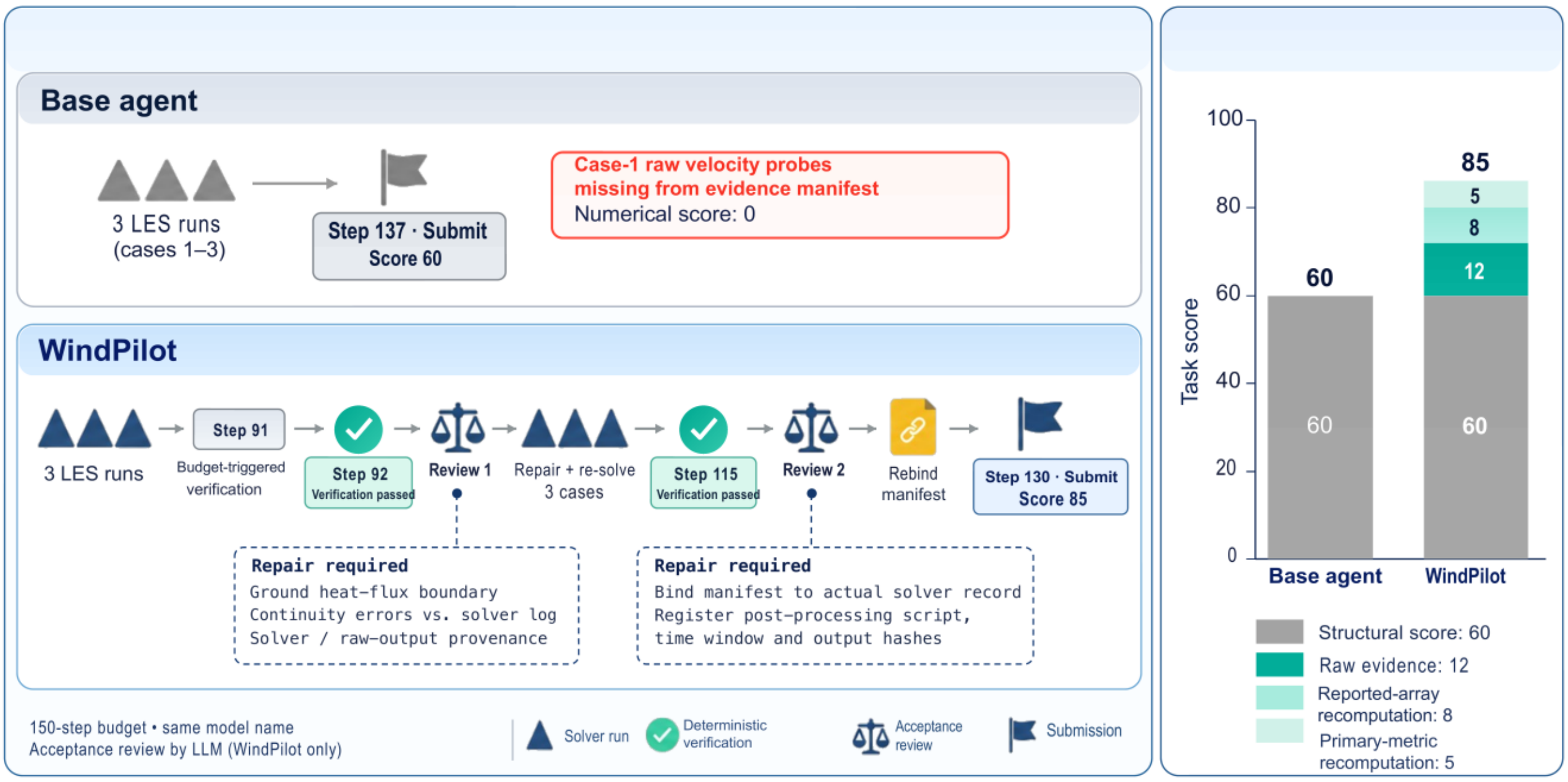

**Fig. 12.** Execution traces for task-094. Nodes follow the recorded action steps; distances between nodes do not represent elapsed time.

The wind-direction rotation and vector-transformation task, task-068, requires transformation of velocity, Reynolds stress and heat flux between geographical wind directions and solver coordinates, with inlet verification through solution. The contract specifies endTime=400 and writeInterval=20 in controlDict, a fixedProfile inlet velocity referencing the designated profile file, and raw sampling at 20 output times and 52 probes.

A budget-triggered check at step 91 located the missing time axis (Fig. 13). Velocity arrays had shape 52×3 rather than raw 20×52×3, while TKE and specific-dissipation-rate arrays had shape 52 rather than 20×52. Stale raw-file/controlDict hashes, an unreadable run manifest and other defects left nine checks failing. The dimensions identified lost sampling times; file mismatches identified broken source relationships.

After re-exporting arrays from raw samples and updating evidence references, all nine checks passed at step 99; checks at steps 116 and 120 found no recurrence. Acceptance review additionally identified incomplete records of actual boundaries, commands and working directories, software versions and executable paths, as well as contract-inconsistent hard-threshold pass rates and counts. The agent addressed these issues and submitted at step 120, scoring 90. The base agent exhausted 150 steps and scored 30, with inconsistent case IDs and missing solver evidence.

Repair in task-068 mainly reorganised existing samples, whereas task-094 required new solves.

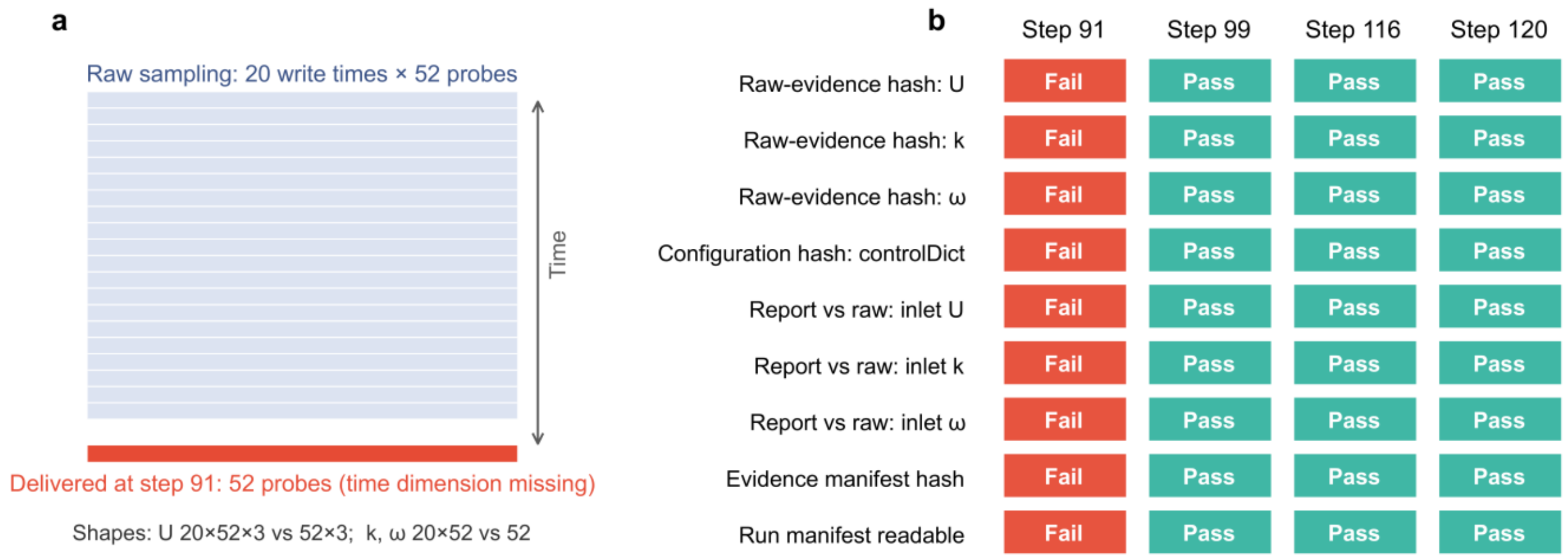

**Fig. 13.** Consistency checks between delivered arrays and raw samples in task-068
(a) Dimensions of raw samples and arrays delivered at step 91. (b) Results of 9 checks across 4 deterministic-verification events.

### *6.6 Verification triggers, error types and final states*

Across 30 tasks, 95 deterministic-verification events were recorded: 81 completion-triggered, 14 budget-triggered and 68 after step 90 (Fig. 14a). Thirteen events detected issues in 10 tasks. First detections were budget-triggered for 8 tasks and completion-triggered for 2.

File accessibility and report–data correspondence dominated the 101 deduplicated errors (Section 5.5; Fig. 14b). Counts were 62 missing/unreadable files, 15 raw-data mismatches, 14 stale hashes, 7 incomplete decision notes, 2 public-threshold exceedances and 1 invalid metric. The first two categories total 77.

Of 101 deduplicated errors, 97 were absent from their tasks' final verification and were therefore counted as repaired (96.0%) under the definition in Section 5.5. All file-access, raw/report, threshold and invalid-metric errors disappeared. Remaining errors were 3 stale hashes and 1 incomplete decision. Figure 14b shows detected/final states by category.

Acceptance review was called 57 times, averaging 1.90 per task: 34 outcomes required repair, 14 indicated insufficient evidence and 9 allowed submission. Reviews followed programmatic passes and thus additionally addressed boundary semantics, run provenance and calculation definitions. Task-094 and task-068 link such issues to subsequent actions.

Cumulative programmatic-verification time averaged 20.8 s per task, with a median of 8.55 s and a maximum of 113 s.

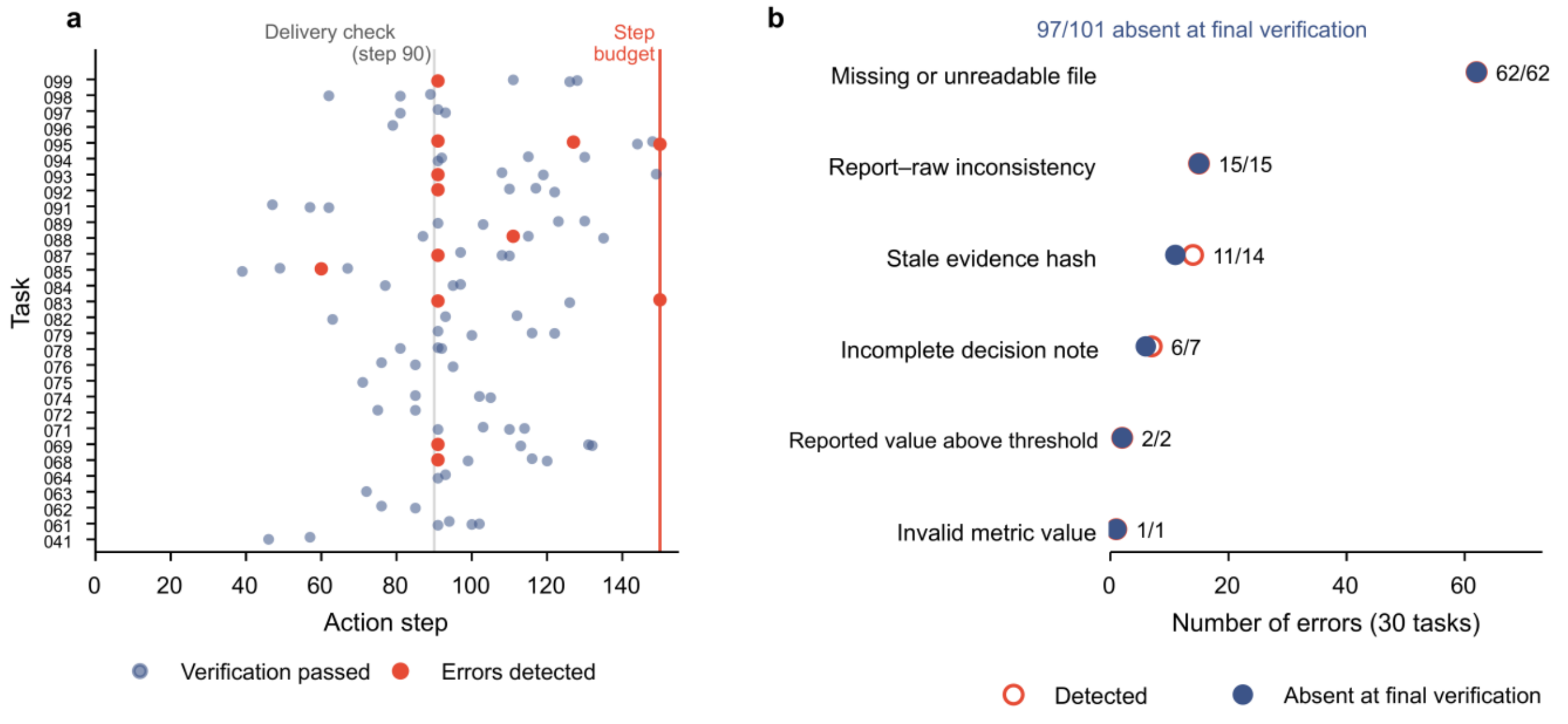

**Fig. 14.** Timing of deterministic verification and detected errors

(a) Positions of 95 deterministic-verification events within task execution; red points indicate detected problems. (b) Detected errors by category (open markers) and errors absent from the final verification (filled markers), defined as repaired in Section 5.5.

# 7 Discussion

## *7.1 Evaluating recomputable deliverables*

A completed simulation run did not by itself provide a usable urban wind result: in most base-agent tasks, deliverables and run records were complete, whereas metrics could not be recomputed from raw samples. Recomputable delivery preserves relationships through the full urban wind calculation. Solver fields become sampled statistics, task metrics and cross-case conclusions. Fixed conditions, probes and windows constrain each step; run records and evidence references identify its computational source.

Structural checks and numerical recomputation have distinct roles. The former establish access to files, runs and evidence; the latter recompute metrics from corresponding outputs and compare them with references. CFD verification and validation address numerical solution error, model representation and external validation [16,17,51]. The task scores defined here measure deliverable organisation, metric recomputation and agreement with designated references; component scores identify what the total has checked.

Contracts preserve meaning across both computations and data processing. Task-093 holds inlet speed, direction, mesh and turbulence model fixed while roughness varies. Task-068 holds the time, point and component relationship between raw samples and arrays.

For an engineering reviewer, provenance identifies which outputs, time span and processing produced a reported value. Recovering the time axis in task-068 restores array–sample correspondence; rebinding runs and processing after boundary changes in task-094 identifies the updated computation.

The evaluation can therefore complement established urban wind checks. Configuration constraints preserve comparison assumptions; recomputation checks statistical processing; references test task-specific results; and observations supply external physical validation [3–5,16,17].

### *7.2 Complementary roles of execution frameworks and domain models*

Execution design can operate outside the underlying LLM. WindPilot keeps constraints accessible, exposes settings and raw outputs through tools, locates findings by files and fields, and lets completion feedback initiate repair. SWE-agent similarly treats the interaction interface as part of execution performance [56]. Urban wind adds the need to maintain computational settings and statistical conventions through those interfaces.

With the same model name and budget, full WindPilot gains 10.0 mean points while reducing mean actions and tokens. Execution configuration therefore contributes to the observed delivery performance.

Improving domain models and organising execution are compatible approaches. CFD-copilot combines adaptation, multi-agent correction and MCP [78], and Foam-Agent/ChatCFD organise configuration, execution and feedback [32,33]. WindPilot examines cross-case urban wind constraints, raw sampling and recomputation under one model name, so model and workflow improvements remain separate opportunities.

Programmatic verification and review support different decisions within the loop. Dimensions, hashes and public metric definitions allow deterministic checking; task semantics and evidence support across files call for review. Task-094 passed programs before review identified boundaries and provenance that led to reruns and rebinding. Task-068 instead began with an array mismatch and then completed run and threshold explanations.

### *7.3 Task-dependent gains and applications in urban wind engineering*

ABL gains have concrete verification targets. Profiles, direction transformations and boundaries expose coordinates, units and array relationships that can be checked against inlet probes and configuration fields. Task-061 demonstrates complete numerical checking from raw evidence to references, and task-068 shows how dimension recovery locates an organisational fault.

All evaluated ABL tasks are D4 and RANS/LES tasks D5, so task type, domain difficulty and solution process vary together across groups. Group results therefore describe these engineering task types under the current configuration.

RANS/LES adds a dependency between settings and time statistics. Non-target conditions must match before contract-defined windows produce comparison metrics. Delivery organisation affects checkability and numerical work affects reference agreement. In task-094, three boundary-adjusted cases were solved again, requiring configuration and subsequent provenance to be updated together.

Checks should follow this engineering order. Comparing settings before solver launch protects the multi-case assumptions; progressively linking outputs and reports during post-processing makes completed runs checkable.

This evaluation covers meshing, ABL and OpenFOAM RANS/LES, which provide basic computations for later diagnosis, risk mapping and design comparison. Other benchmark families retain PALM, GIS and decision-making requirements. Extensions can reuse the organisation of contracts, runs, raw data and metrics while developing quantity-specific checks and references.

*7.4 Verification strategies and application boundaries for long solver runs*

For long solves, the timing of checks matters because later responses are costly. Budget-triggered checks detected most first-recorded problems before completion, leaving room for organisation and repair. Completion checks then examined the files offered for submission.

The appropriate moment depends on the checked object. Settings and fixed conditions can be inspected before solution, arrays and time dimensions as outputs and statistics appear, and provenance and metric references whenever delivery changes. Re-export in task-068 and adjusted reruns in task-094 illustrate different responses.

Verification and the responses it triggers have different resource costs. Verification averages 20.8 s per task, chiefly checking files, arrays and public rules. Review uses model calls, re-export uses existing samples and reruns add numerical computation.

Verification rules should expand along metric provenance. Public thresholds, raw logs, probe arrays and post-processing definitions have distinct access points. Associating these with task fields can progressively extend the programmatically checkable objects while existing contracts, evidence references and run records are retained.

## 8 Conclusions

For the evaluated urban wind CFD tasks, a completed simulation run was not sufficient for a usable result: at least the structural maximum of 60 points was reached by the base agent on 28 of 30 tasks, but recomputable metrics were delivered on only 6. UrbanWind-AgentBench organises 200 tasks and 367 cases across 10 urban wind engineering families. Public contracts connect changed factors, fixed conditions, sampling and delivery requirements; structural checks and numerical recomputation evaluate the results.

On 30 independently referenced OpenFOAM tasks, full WindPilot raised mean scores from 62.5 to 72.5 with the same model name and budget. Mean action steps fell 19.5%, mean tokens 12.9%, and ABL scores rose 18.5 points. Components and traces connect these changes to raw-evidence access, array recomputation, boundary corrections and run-record binding.

Task constraints, provenance and delivery checks are thus supported as components of urban wind agent execution frameworks. Multi-case automation must preserve comparison conditions while tying the report to actual solves and samples. WindPilot joins runtime feedback with independent scoring after submission.

## Data and code availability

The project website is available at https://ttanddd.github.io/urbanwind-agentbench/. Task definitions, public inputs, the evaluation framework, the WindPilot harness and supporting data are available at https://github.com/TTANDDD/urbanwind-agentbench. Reference coverage and release status are documented in the repository.